\documentclass[runningheads]{llncs}

\usepackage{eccv}

\usepackage{eccvabbrv}

\usepackage{threeparttable}

\usepackage{graphicx}
\usepackage{booktabs}
\usepackage{multirow}%

\usepackage{mathrsfs}%
\usepackage[title]{appendix}%
\usepackage{xcolor}%
\usepackage{textcomp}%
\usepackage{manyfoot}%
\usepackage{booktabs}%
\usepackage{listings}%

\usepackage{microtype}
\usepackage{adjustbox}
\usepackage{hhline}
\usepackage{colortbl}
\usepackage{tabularray}
\usepackage{physics}
\usepackage{hyperref}
\usepackage{url}
\usepackage{nicefrac}       
\usepackage{pifont}
\usepackage{mdframed}
\usepackage{fontawesome}
\usepackage{mathtools}

\usepackage[accsupp]{axessibility}  

\usepackage{hyperref}
\usepackage{wrapfig}
\usepackage{subcaption}
\usepackage{caption}
\usepackage[linesnumbered,ruled]{algorithm2e}

\newcommand{\figref}[1]{Fig.~\ref{#1}}
\newcommand{\secref}[1]{Sec.~\ref{#1}}

\begin{document}

\title{MVGS: Multi-view Regulated Gaussian Splatting  \\ for Novel View Synthesis}

\titlerunning{MVGS: Multi-view Regulated Learning}

\author{Xiaobiao Du\inst{1, 2} \and
Yida Wang\inst{2} \and
Xin Yu\inst{3}\thanks{Corresponding author: xin.yu@adelaide.edu.au}}

\authorrunning{Du et al.}

\institute{University of Technology Sydney \and
 Li Auto Inc. \and
Australian Institute for Machine Learning,  Adelaide University \\
}

\maketitle

\begin{abstract}
     Recent works in novel view synthesis, \textit{e.g.}, Neural Radiance Field (NeRF) and 3D Gaussian Splatting (3DGS), have significantly advanced rendering quality and efficiency. 
However, existing Gaussian-based novel view synthesis methods typically follow a single-view optimization paradigm.
We observed that this optimization paradigm suffers from unstable gradients, leading to suboptimal rendering quality.
To tackle this issue, we present a novel multi-view regulated Gaussian Splatting (MVGS) that fully leverages a multi-view coherent (MVC) constraint throughout the optimization process. 
Specifically, our proposed MVC enhances 3D Gaussian multi-view consistency and thus ensures smoother gradient updates. 
Furthermore, since single-scale training usually leads to suboptimal solutions, we propose a cross-intrinsic guidance scheme in a coarse-to-fine manner to improve the convergence of multi-view optimization in 3DGS. 
In particular, by incorporating more multi-view images at the low resolution, we can optimize 3D Gaussians with more comprehensive perspectives. Then, finer-scale Gaussians are initialized by coarsely estimated ones instead of optimizing full-scale 3D Gaussians from scratch.
Moreover, we found that 3D Gaussians usually struggle to fit 2D training views with minimal overlap.
Thus, we propose a novel multi-view cross-ray densification strategy, where 3D Gaussians are dynamically split to accommodate drastic viewpoint variations in the multi-view optimization process.
In this way, the multi-view consistency can be further improved. 
Notably, our proposed MVGS method is a plug-and-play optimizer. 
Extensive experiments across various tasks demonstrate that our proposed MVGS improves existing Gaussian-based methods and achieves state-of-the-art performance. 
Project Page: \href{https://xiaobiaodu.github.io/mvgs-project/}{\color{magenta}{https://xiaobiaodu.github.io/mvgs-project/}}
\end{abstract}

\section{Introduction}

Photorealistic rendering of unbounded scenes or objects holds considerable significance in both industry and academia, \textit{e.g.} multi-media generation~\cite{du2024ethics, guo2024being, shen2023auslan, shen2024mm,shen2025cross, shenbanz, shen2025fingercap}, virtual reality~\cite{vrgs}, human body understanding~\cite{guo2025plnet, guo2026beyond}, and autonomous driving~\cite{du2024dreamcar, du20243drealcar}. 
Conventional primitive-based representations based on mesh and point clouds~\cite{botsch2005high, lassner2021pulsar, yifan2019differentiable, munkberg2022extracting}, leverage efficient rasterization to achieve real-time rendering.
Although these methods deliver high efficiency, they still struggle to reconstruct fine-grained and precise appearance, leading to blurry artifacts and discontinuity.
On the contrary, implicit representation~\cite{erler2020points2surf} and neural radiance field~\cite{mildenhall2021nerf, mip-nerf360, instant-ngp}(NeRF), employ the multi-layer perceptron (MLP) to improve the fidelity of scene geometry, thus attaining high-quality geometric and appearance details. 
However, their inference efficiency is still limited, even when employing acceleration operators, like Instant-NGP~\cite{muller2022instant}.

\begin{figure*}[t!]
    \centering
    \includegraphics[width=0.9\linewidth]{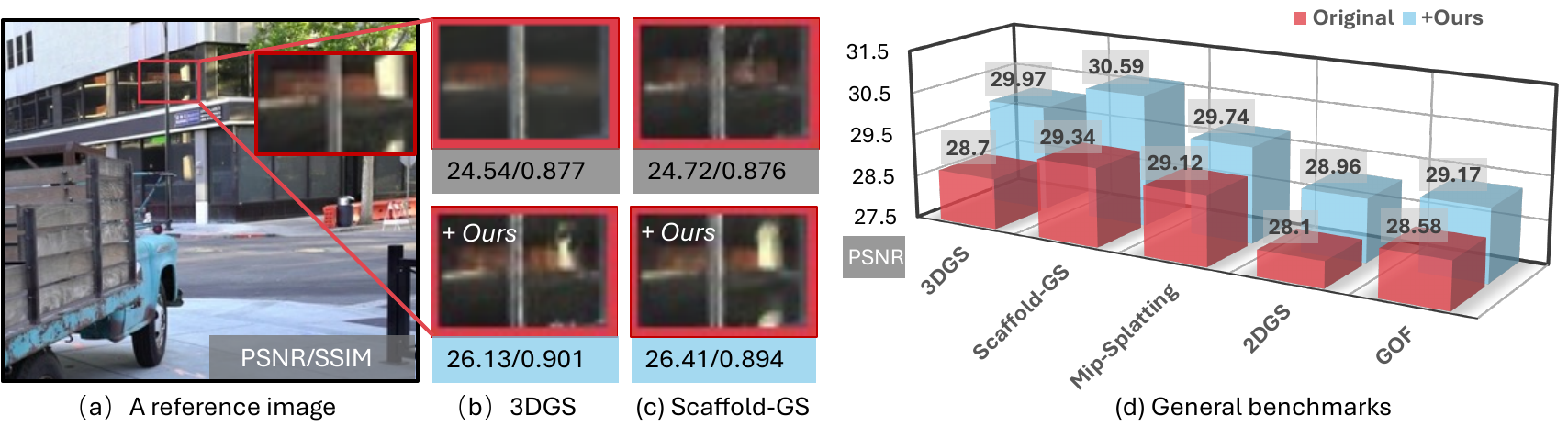} 
    \caption{\emph{\textbf{MVGS}} supplements general improvements for Novel View Synthesis for Gaussian-based methods~\cite{kerbl2023gaussiansplatting, lu2024scaffold}, as shown in \textbf{(b)} and \textbf{(c)}. Extensive experiments are conducted to demonstrate that our proposed method delivers consistent improvements in \textbf{(d)}.
    }
    \label{fig:teaser}
    
\end{figure*}


Recently, 3D Gaussian-based explicit representations, \textit{e.g.}, Gaussian Splatting (3DGS)~\cite{kerbl2023gaussiansplatting,jiang2024gaussianshader,lu2024scaffold,4dgs}, achieve both state-of-the-art rendering quality and efficiency.
Specifically, 3DGS is firstly initialized with point clouds which are either scanned or learned from Structure from Motion (SfM)~\cite{sfm2016}, and then a set of 3D Gaussian kernels are optimized to represent a whole scene with realistic appearances.
However, existing Gaussian-based methods usually employ a single-view training paradigm, \textit{i.e.}, training with only a single camera view per iteration.
We observed that this single-view training paradigm may lead to unstable gradients in updating 3D Gaussian kernels, ultimately resulting in unsatisfactory results.

In this paper, we propose a novel multi-view regulated Gaussian Splatting optimization method, dubbed MVGS. Our MVGS is a generic optimization method compatible with various Gaussian-based approaches, and it can also further improve Novel View Synthesis (NVS) performance, as shown in \figref{fig:teaser}. 
Unlike prior arts, our MVGS incorporates multi-view images in each optimization step to regulate multi-view learning.
Specifically, we impose multi-view consistency constraints when learning the 3D Gaussian parameters. In this manner, one 3D Gaussian is jointly updated by multi-view images, thus significantly smoothing its gradient descent process.


To facilitate the convergence of multi-view optimization, we propose a cross-intrinsic guidance scheme in training from low-resolution to high-resolution.
The low-resolution training is designed to capture multi-view global and coarse appearances rapidly. Then, we focus on refining 3D details more efficiently in high-resolution training. Note that, in the low-resolution training phase, we employ more multi-view images in optimization as the image resolution is much smaller. As a result, the optimization in the low-resolution phase provides a more holistic yet better initialization for the high-resolution phase.
In addition, we found that limited 3D Gaussians often struggle to fit drastic viewpoint variation.
Hence, we propose a multi-view cross-ray densification strategy that consists of multi-view augmented densification and cross-ray densification.
To be specific, the proposed multi-view augmented densification is designed to enhance 3D Gaussians when input views exhibit minimal overlaps, thereby facilitating both multi-view training and fine detail reconstruction.
Meanwhile, the cross-ray densification is proposed to densify 3D Gaussians in the overlapped 3D space where rays emitted from multiple views intersect.
With this strategy, more 3D Gaussians can be enhanced to better fit drastic view changes, and the multi-view consistency can be further enhanced.

Extensive experiments are conducted to demonstrate that our method significantly improves NVS performance for existing Gaussian-based methods across various tasks, including general and reflective object NVS, dynamic and large-scale scene Novel View Synthesis.
Notably, our results reveal that appropriately increasing the number of multi-view inputs in each optimization round leads to a corresponding improvement in Novel View Synthesis.
In conclusion, we summarize our contributions as below:

$\bullet$ We propose a plug-and-play multi-view regulated Gaussian Splatting (MVGS) method that imposes multi-view coherent constraints to smooth gradient updates for better NVS performance.


$\bullet$ We propose a multi-view cross-ray densification strategy to derive sufficient 3D Gaussians for better fitting drastically varied viewpoints. In this way, we significantly mitigate the viewpoint discrepancy issue in the multi-view optimization and facilitate multi-view learning.

$\bullet$ Extensive experiments demonstrate that our method is a universal optimization solution, achieving obvious performance improvement for existing Gaussian-based methods across various tasks, including static object, scene, and dynamic 4D NVS.

\section{Related Work}
In this section, we first review recent volume rendering techniques used by the Neural Radiance Field, which have significantly advanced 3D reconstruction and rendering. Then, we discuss recent developments in Gaussian-based explicit representation models, including 3DGS and its variants that are related to our work.

\noindent \textbf{Neural Field Rendering:}
Significant advancements have been made in novel-view synthesis, particularly since the introduction of NeRF~\cite{mip-nerf, mildenhall2021nerf}, which employs MLPs to parameterize geometry and view-dependent appearance through an implicitly defined radiance field. 
Moreover, the training and inference efficiency of NeRF has been enhanced with hash-grid~\cite{instant-ngp} and explicitly defined samplers~\cite{li2023nerfacc}. 
Ref-NeRF~\cite{refnerf} introduces a parametric representation of reflected radiance and structure with spatially varying scene properties.
This representation significantly improves the accuracy and realism of specular reflections.
TensoRF~\cite{TensoRF} proposes a 3D voxel grid representation with multiple channels. The proposed representation with compact low-rank tensor components speeds up the rendering efficiency.
Built on top of the radiance field, NeuS~\cite{wang2021neus}, NeuS2~\cite{wang2022neus2}, and HF-NeuS~\cite{wang2022hf} also perform more precise surface reconstruction against traditional MVS fusion, such as MeshMVS~\cite{shrestha2021meshmvs}. Given all the advantages of neural rendering, its efficiency is still not satisfactory.

\noindent \textbf{Gaussian Splatting:}
Recently, 3D Gaussian Splatting (3DGS)~\cite{kerbl2023gaussiansplatting} has been proposed, demonstrating impressive real-time NVS performance. 
3DGS rasterizes 3D Gaussian spheres that are projected through $\alpha$-blending and depth-sorting, and achieves real-time rendering efficiency by avoiding complex ray tracing.
Thanks to its real-time rendering speed and high-quality reconstruction performance, 3DGS has been applied to numerous tasks, such as autonomous driving~\cite{yan2024streetgs}, reflective object reconstruction~\cite{jiang2024gaussianshader}, and 4D reconstruction~\cite{4dgs}.
Subsequent works focus on improving Gaussian representations, such as 2D Gaussian Splatting (2DGS)~\cite{2dgs} and structure grid representations~\cite{lu2024scaffold}. 
GaussianPro~\cite{cheng2024gaussianpro} proposes a normal propagation method to bridge the gap from SfM initialization and mitigate densification limitations. 
Pixel-GS~\cite{pixelgs} proposes a gradient-based scaling densification strategy to avoid the generation of floaters near the camera.
PixelSplat~\cite{charatan2024pixelsplat} predicts a dense probability distribution over sampled 3D Gaussian positions.
Mobile-GS~\cite{du2026mobile} is the first one to deploy 3DGS on mobiles and achieves real-time rendering, but it still follows the original single view iterative training of 3DGS.
3DGS-MCMC~\cite{mcmc} proposes a Markov Chain Monte Carlo (MCMC) method, where Gaussian kernels represent samples in a probabilistic scene representation. 
This approach enables efficient rendering by leveraging the stochastic nature of MCMC to approximate complex 3D structures and radiance fields.
However, these Gaussian-based explicit representation methods adopt a single-view optimization strategy~\cite{mallick2024taming, rota2024revising, kerbl2023gaussiansplatting}, leading to unstable gradients in training and unsatisfactory results in novel view synthesis.
Some works have been proposed to utilize multi-view features~\cite{chen2024mvsplat, wewer2024latentsplat}.
Built on pre-trained networks, the extracted multi-view features can solve some of the aforementioned difficulties. For example, MVSplat~\cite{chen2024mvsplat} builds a cost volume representation to store cross-view similarities for depth estimation.
LatentSplat~\cite{wewer2024latentsplat} proposes a representation encoding uncertainty with latent Gaussian features.
AbsGS~\cite{ye2024absgs} analyzes the cause of floaters and proposes to use the view-space positional gradient as guidance for densification.
Grendel~\cite{zhao2024grendel} proposes a distributed GPU strategy to speed up the GS training process, but ignores the multi-view training for enhancing multi-view consistency. 
In this paper, our proposed method provides a more general solution that is compatible with many 3DGS variants and does not rely on pre-trained models to impose multi-view constraints.

\section{Methodology}
\label{sec:method}

3D Gaussian Splatting (3DGS) represents scenes using explicit anisotropic 3D Gaussians $\mathcal{G} = \{\boldsymbol{\theta}_i\}_{i=1}^N$ where each Gaussian $\boldsymbol{\theta}_i = (\mu_i, \Sigma_i, c_i, o_i)$ is parameterized by position $\mu_i \in \mathbb{R}^3$ (mean) in world coordinates, covariance $\Sigma_i \in \mathbb{S}_{++}^3$ (positive definite matrix) controlling rotation and scale,
view-dependent color $c_i \in \mathbb{R}^3$ via spherical harmonics basis $\psi(\omega):\mathbb{S}^2 \to \mathbb{R}^3$ where $\omega$ is view direction, and opacity $o_i \in [0,1]$ controlling light transmission.
The differential rendering equation accumulates 3D Gaussian contributions by the alpha blending:
\begin{align}
    \mathbf{C}&= \sum_{i=1}^N c_i \alpha_i \prod_{j=1}^{i-1}(1-\alpha_j), \\
    \alpha_i &= o_i \exp\left(-\frac{1}{2}\Delta x_i^T\Sigma_i^{-1}\Delta x_i\right),
\end{align}
where $\Delta x_i = x_i-\mu_i$ denotes the positional offset from the Gaussian position to the sampled point $x_i$. 
This explicit representation enables real-time rendering through tile-based rasterization with $\mathcal{O}(N)$ complexity, without considering all Gaussian attributes for simplicity. However, the single-view training optimization strategy adopted by existing Gaussian-based methods introduces unstable gradients in optimization, leading to inferior performance.
Hence, we propose to improve the performance of 3DGS by an innovative multi-view constraint in \secref{sec:mvgs}, enhance multi-scale features by enriching different intrinsic setups in \secref{sec:intrinsic}, and address the minimal overlap among multi-view inputs through densification in \secref{sec:crd}.

\begin{figure*}[t!]
    \centering
    \includegraphics[width=0.93 \linewidth]{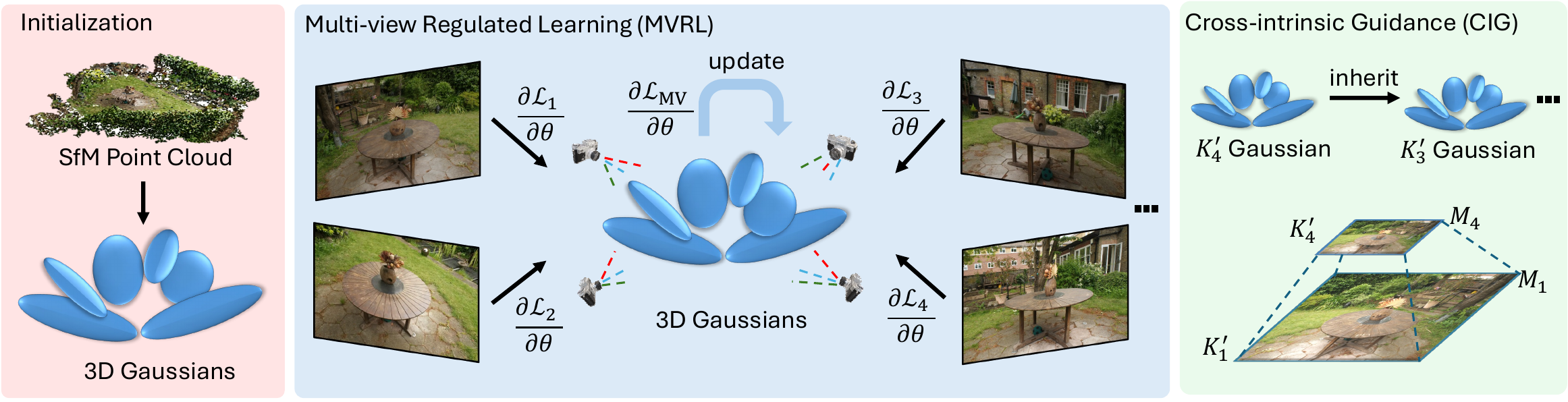} 
    \caption{\textbf{Illustration of our proposed \emph{MVGS}.}
    The proposed multi-view regulated training imposes multi-view constraints in learning 3D Gaussians. These constraints encourage the optimization of 3D Gaussians to align with diverse viewpoints, leading to improved reconstruction quality. Additionally, cross-intrinsic guidance further enhances multi-view learning across different scales.
    }
    \label{fig:method}
\end{figure*}

\subsection{Multi-View Regulated Training}
\label{sec:mvgs}
3DGS is supervised by a single-view image per iteration,
where the supervision viewpoint is randomly selected for each iteration.
The loss function of the original 3DGS can be formulated as:
\begin{equation}
\begin{split}
     \mathcal{L} &=  (1-\lambda) \mathcal{L}_1
      + \lambda \mathcal{L}_\text{D-SSIM},
\end{split}
\end{equation}
where $\mathcal{L}_1$ and $\mathcal{L}_\text{D-SSIM}$ denote the mean absolute error and D-SSIM loss~\cite{kerbl2023gaussiansplatting}, respectively.
The hyperparameter $\lambda$ balances the contributions of these two loss terms.
Note that in the single-view supervision fashion, only partial 3D Gaussians would be updated by the gradient descent.
Moreover, some Gaussians that satisfy the single-view image supervision may conflict with physical restrictions, potentially leading to unstable gradients. 

In this work, we propose a multi-view regulated training with multi-view gradient aggregation.
For Gaussian primitive parameters $\boldsymbol{\boldsymbol{\boldsymbol{\theta}}} \in \{\mu_i, \Sigma_i, o_i, c_i\}$, a complete gradient chain is computed through:
\begin{equation}
    \frac{\partial\mathcal{L}_{\text{MV}}}{\partial\boldsymbol{\boldsymbol{\theta}}} = \sum_{m=1}^M\frac{\partial\mathcal{L}_m}{\partial\boldsymbol{\boldsymbol{\theta}}},
    \label{equ:gradient_chain}
\end{equation}
where $\mathcal{L}_m$ denotes the loss $\mathcal{L}$ under the $m$-th view.
Here, we do not average the gradient but sum it to obtain a larger gradient magnitude.
Note that summing gradients is equivalent to optimizing the summed multi-view loss, and its effect can be interpreted as adaptively increasing the effective learning rate for Gaussians jointly observed by multiple views.
In this way, 3D Gaussians supervised jointly by multiple views are optimized in a more holistic and comprehensive way, leading to stable gradients and better performance. 
To better understand the benefits of multi-view regulated training, we provide a theoretical analysis of its gradient variance and directional properties.

\begin{figure}[t!]
    \centering
    \includegraphics[width=0.7\linewidth]{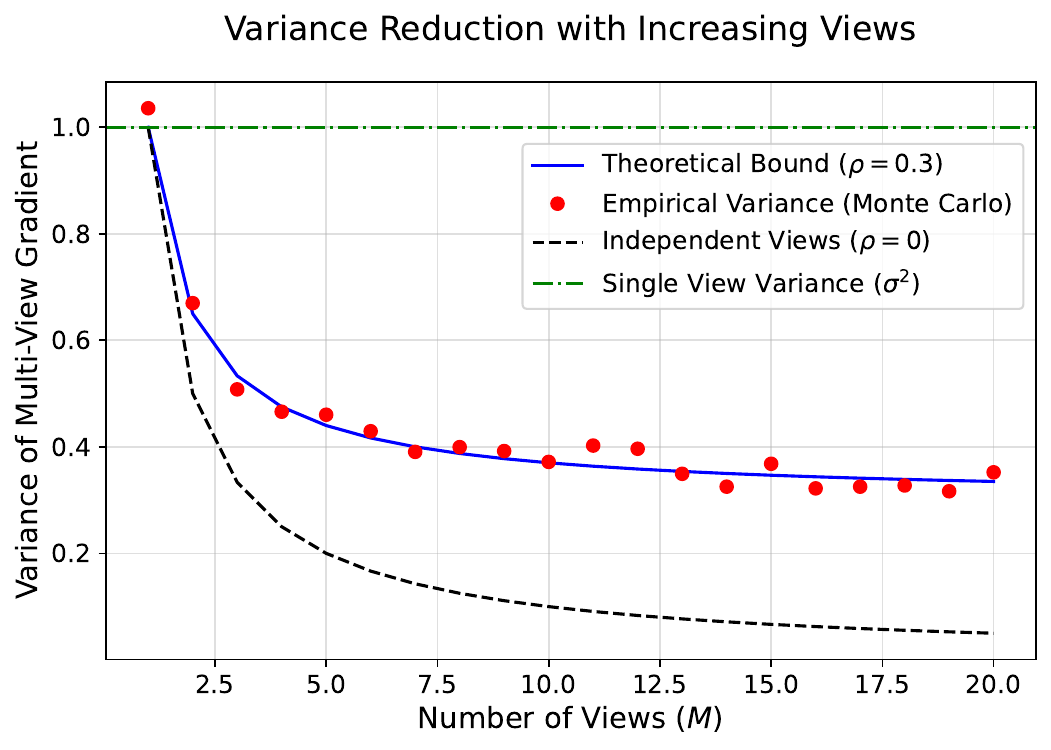}
    \caption{Relative Variance reduction with increasing views $M$ (simulated with $\rho = 0.3$). Multi-view gradients (blue) show faster variance decay compared to independent gradients (red dashed).}
    \label{fig:variance}
\end{figure}

\noindent \textbf{Variance Bound Derivation:}
We analyze the variance of the multi-view gradients by considering the properties of the variance of randomly selected input views. 
The variance of our multi-view loss $\mathrm{Var}\left[\frac{\partial\mathcal{L}_{\text{MV}}}{\partial\boldsymbol{\boldsymbol{\theta}}}\right]$ can be formulated as:
\begin{equation}
\sum_{m=1}^M \mathrm{Var}\left[\frac{\partial\mathcal{L}_m}{\partial\boldsymbol{\boldsymbol{\theta}}}\right] + 2\sum_{1\leq i<j\leq M}\mathrm{Cov}\left[\frac{\partial\mathcal{L}_i}{\partial\boldsymbol{\boldsymbol{\theta}}}, \frac{\partial\mathcal{L}_j}{\partial\boldsymbol{\boldsymbol{\theta}}}\right].
\end{equation}

Here, we sample independently and identically distributed view gradients in single-view systems \cite{tsai2016multiview}:
   $ \mathrm{Var}[\frac{\partial\mathcal{L}_m}{\partial\boldsymbol{\boldsymbol{\theta}}}] = \sigma^2,~~ \forall m$, and $\sigma^2$ is the single-view gradient variance.
   For simplicity, we discuss the case of independently and identically distributed view gradients in single-view systems. 
   Thus, under a uniform pairwise correlation assumption, the variance can be expressed as:
\begin{equation}
    \mathrm{Var}\left[\frac{\partial\mathcal{L}_{\text{MV}}}{\partial\boldsymbol{\boldsymbol{\theta}}}\right] \leq M\sigma^2 + 2\sum_{1\leq i<j\leq M}\mathrm{Tr} \left[\mathrm{Cov}[\frac{\partial\mathcal{L}_i}{\partial\boldsymbol{\boldsymbol{\theta}}}, \frac{\partial\mathcal{L}_j}{\partial\boldsymbol{\theta}}]\right].
\end{equation}

We denote $\rho = \frac{1}{\sigma^2}\mathbb{E}[ \mathrm{Tr} \left( \mathrm{Cov}[\frac{\partial\mathcal{L}_i}{\partial\boldsymbol{\theta}}, \frac{\partial\mathcal{L}_j}{\partial\boldsymbol{\theta}}]\right) ]$
be the average correlation coefficient. For $M$ views, 
we consider the general case where gradients from different views are not fully correlated, leading to a bounded variance:
\begin{equation}
    \mathrm{Var}\left[\frac{\partial\mathcal{L}_{\text{MV}}}{\partial\boldsymbol{\theta}}\right] \leq M\sigma^2\left(1 + (M-1)\rho\right),
\end{equation}
where the multi-view aggregation preserves directional information while suppressing noise. The relative variance reduction is empirically validated in \figref{fig:variance}, showing faster decay with increasing $M$ compared to independent gradients.

\noindent \textbf{Directional Consistency:}
The directional consistency of multi-view gradients emerges from the inner product:
\begin{equation}
    \left\|\frac{\partial\mathcal{L}_{\text{MV}}}{\partial\boldsymbol{\theta}}\right\|_2^2 = \sum_{m=1}^M\left\|\frac{\partial\mathcal{L}_m}{\partial\boldsymbol{\theta}}\right\|_2^2 + 2\sum_{i<j}\left\langle\frac{\partial\mathcal{L}_i}{\partial\boldsymbol{\theta}},\frac{\partial\mathcal{L}_j}{\partial\boldsymbol{\theta}}\right\rangle,
\end{equation}
where this equation decomposes the squared gradient magnitude into two components. The first term, $\sum_{m=1}^M\|\frac{\partial\mathcal{L}_m}{\partial\boldsymbol{\theta}}\|_2^2$, sums the squared gradient magnitudes of each view, representing the total energy of the gradients without considering their directions. The second term, $2\sum_{i<j}\langle\frac{\partial\mathcal{L}_i}{\partial\boldsymbol{\theta}}, \frac{\partial\mathcal{L}_j}{\partial\boldsymbol{\theta}}\rangle$, captures the similarity between view gradients.
This similarity acts as a regulation term that controls how gradient directions from different views influence the final update. When gradients are coherent across views, their inner products become positive, amplifying the overall update signal. In contrast, incoherent or conflicting gradients lead to reduced magnitude, thereby suppressing unstable or inconsistent updates.
By aggregating gradients across multiple viewpoints, the model is effectively regularized toward consistent geometric structure.

 Multi-view gradients encoding similar structural information enhance geometric consistency, such as surface normals and depth, while exhibiting discrepancies in view-dependent effects, such as specular highlights. This property contains the following benefits:

\begin{figure}[t!]
    \centering
    \includegraphics[width=0.99\linewidth]{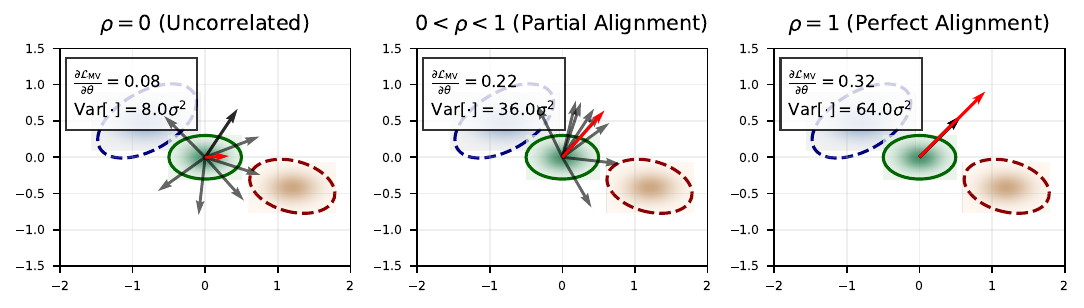}
    \caption{\textbf{Toy example of using three Gaussian primitives to investigate $\rho$.} The black arrow denotes the camera observation direction. The red arrow indicates the direction of the final gradient update. We illustrate different correlation states of $\rho$, ranging from uncorrelated to perfectly correlated.} 
    \label{fig:correlation_analysis}
\end{figure}

\noindent\textbf{Uncorrelated Gradients (\( \rho \approx 0 \)): } 
When gradients are uncorrelated, the standard deviation of the aggregated gradient scales as \(\mathcal{O}(\sqrt{M})\), while the mean magnitude scales as \(\mathcal{O}(M)\), leading to improved signal-to-noise ratio.
It indicates effective noise suppression, reducing noise and stabilizing training.  When more views are incorporated, uncorrelated noise would be suppressed more significantly.

\noindent\textbf{Highly Correlated Gradients (\( \rho \approx 1 \)):}
When gradients are highly correlated, the total gradient magnitude scales as \( M^2\sigma^2 \), suggesting a strong coherent signal across views. This correlation reflects substantial structural consistency in the scene, leading to an 
\(\mathcal{O}(M)\) increase in gradient magnitude. As a result, parameter updates for 3D Gaussians become more reliable, reinforcing structural information during optimization.

\noindent\textbf{Intermediate Correlation (\( 0 < \rho < 1 \)):}
 Partially correlated gradients (\( 0 < \rho < 1 \)) interpolate between \(\mathcal{O}(\sqrt{M})\) and \(\mathcal{O}(M)\).
This balance allows the model to leverage multi-view consistency while mitigating the impact of noise. The increased gradient magnitude facilitates escape from local minima while preserving stable optimization dynamics, potentially improving overall performance.

As shown in Fig.~\ref{fig:correlation_analysis}, the variance scaling law 
$\text{Var}[\frac{\partial\mathcal{L}_{MV}}{\partial\boldsymbol{\theta}}] = M\sigma^2(1 + (M-1)\rho)$  manifests how gradient variance evolves with the number of views.
Overall, our proposed multi-view regulated training enhances the learning process by ensuring that essential scene features are consistently reinforced, while view-dependent noise is minimized, leading to more stable and accurate gradient-based optimization in multi-view settings.

\subsection{Cross-intrinsic Guidance}

\label{sec:intrinsic}

Inspired by the image pyramid, we propose a coarse-to-fine training scheme with different camera setups, \textit{i.e.} intrinsic parameters $K$.
Specifically, a 4-layer image pyramid with downsampling factors $\mathbf{S} = \{2^{k-1} ~|~ k=4, 3, 2, 1\}$ could be constructed by simply reconfiguring the focal length $f$ and principal point $c$ in $K$ through:
\begin{equation}
    K'_s = \begin{bmatrix}
        f/s & 0 & c_x/s \\
        0 & f/s & c_y/s \\
        0 & 0 & 1
    \end{bmatrix}, \quad s \in \mathbf{S}.
\end{equation}

Empirically, the largest downsampling factor $s$ set as 8 is enough to accommodate sufficient training images for multi-view training. The smallest downsampling factor set as 1 means that the downsampling operation is not applied. Note that the excessively large downsampling factor leads to blurry ground truth, thereby decreasing the fidelity of reconstructed 3D Gaussians and increasing the difficulty for later refinement.

For each layer, we incorporate different extents of multi-view consistency with matched multi-view settings $M_s$ = $\{ M_4, M_3, M_2, M_1 \}$. 
In particular, the larger downsampling factor enables more views to provide stronger multi-view constraints. 
In the initial three training stages, we run only a few thousand iterations per stage without completely training the model.
Since target images are downsampled, the model cannot capture sufficient details during these early stages.
Therefore, we treat the first three training stages as coarse training. We train these 3D Gaussians from coarse to fine under the guidance of cross-intrinsics.

During coarse training, incorporating more multi-view information imposes more holistic constraints on the entire 3D Gaussians.
In this case, the rich multi-view information provides thorough supervision for the whole 3DGS and encourages fast fitting with coarse texture and structure. 
Once the coarse training is finished, fine training is started. Thanks to the previous coarse training stages providing a coarse architecture of 3DGS, the fine training stage only needs to refine and sculpt fine details for each 3D Gaussian.
Especially, the coarse training stages provide more powerful multi-view constraints.
It conveys the learned multi-view feature to the next fine training. 
This scheme effectively enhances multi-view constraints and further improves rendering quality.

\begin{figure}[t!]
    \centering
    \includegraphics[width=0.89 \linewidth]{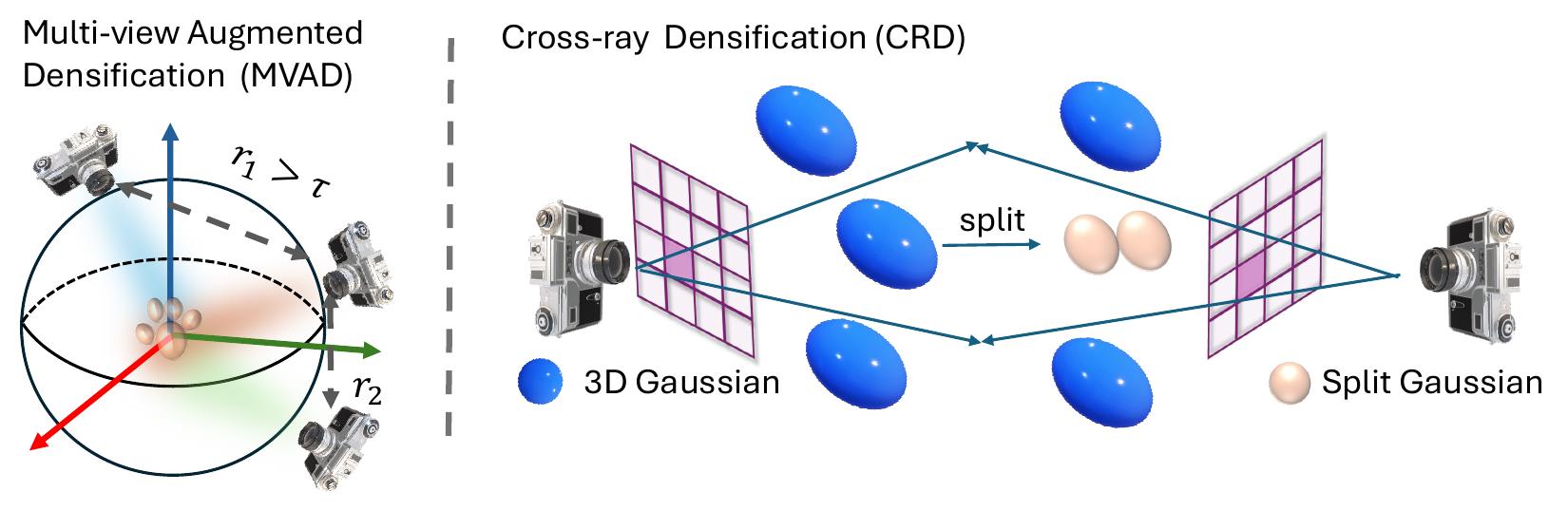} 
    \caption{\textbf{Illustration of the proposed multi-view cross-ray densification strategy.} This strategy is divided into multi-view augmented densification and cross-ray densification. We propose to densify 3D Gaussians responsible for multiple views to a certain volume for finer fitting results and better multi-view consistency.
    }
    \label{fig:crossray}
\end{figure}

\subsection{Multi-view Cross-ray Densification Strategy}
\label{sec:crd}
The proposed multi-view cross-ray densification strategy can be divided into multi-view augmented densification and cross-ray densification. We rely on the proposed multi-view process and ray tracing technique to locate and densify the 3D Gaussians for multi-view consistency enhancement.

\subsubsection{Multi-view Augmented Densification}
To obtain fast convergence and fine-grained Gaussian kernels, we propose a multi-view augmented densification module.
Specifically, the proposed multi-view augmented densification module can control the threshold of the densification effects for 3D Gaussians to make sure multi-view consistency.
As depicted in \figref{fig:crossray}, this module first estimates whether the training views are strongly distinct. 
Instead of using the original camera translations directly, we normalize the camera translations of sampled views into a unit sphere. 
It makes our strategy adaptable to various scenes. 
Then, we measure the multi-view consistency $\{r_i~|~i = 1, 2, \dots, n \}$ by computing the relative translation distances and the similarity of camera rotation matrix between each camera and another, where the number $n$ is $(M^2-M)/2$, given $M$ training views. 
In this way, we can measure the discrepancies from different views with each other.

In our multi-view augmented densification module, we modify the final threshold for enhancing densification via
$\hat{\beta} = \beta - \frac{\beta}{2} H\left(\frac{r_i}{\tau} - 1\right),$
where $H(\cdot)$ is Heaviside function, returning $1$ if the input is larger or equal $0$. 
$\tau$ is a predefined hyperparameter, adjusting the extent of the discrepancy between each camera.
$\beta$ is the original threshold for densification.
When the discrepancies between each view become large, we enhance the densification effect by half the original densification threshold. In this way, we can further facilitate the multi-view training. 
Consequently, our proposed multi-view augmented densification module allows 3D Gaussians to fit better with each view and capture finer scene details.

\subsubsection{Cross-ray Densification}
To enhance multi-view constraints during optimization, we propose a cross-ray densification strategy that selectively increases the density of 3D Gaussians in the multi-view overlapped regions. Due to the explicit nature of 3D Gaussian representations, certain regions have a significant influence on the rendering quality from multiple viewpoints. However, directly identifying these regions in 3D space is challenging.
As illustrated in \figref{fig:crossray}, we propose a cross-ray densification strategy, starting from 2D space and then searching in 3D adaptively. Specifically, we first calculate loss maps of multiple views and then locate the regions containing the largest average loss values using a sliding window with size $(h, w)$. Afterward, we cast rays from the vertices of these regions with four rays per window. Then, we calculate the intersection planes across rays of different perspectives. Since we cast rays per perspective, the intersection planes can form several cuboids. These cuboids are the multi-view overlapped regions containing significant 3D Gaussians that play an important role when rendering for multiple views. Therefore, we densify more 3D Gaussians in these overlapped regions to enhance multi-view constraints and facilitate the training of multi-view supervision.

\begin{table}[t!]
\caption{\textbf{Quantitative results of state-of-the-art novel view synthesis methods on real-world datasets.} We report results on three commonly used datasets, including Mip-NeRF 360~\cite{mip-nerf360}, Tank\&Temples~\cite{tanktemple}, and Deep Blending~\cite{deepblending}.    The \colorbox{red!30}{best}, \colorbox{orange!30}{second best}, and \colorbox{yellow!30}{third best} results are denoted by red, orange, and yellow, respectively. }
\centering
\resizebox{0.95\linewidth}{!}
{
    \begin{tabular}{ l | l l l | l l l| l l l }
    \toprule
    Dataset & \multicolumn{3}{c|}{Mip-NeRF360} & \multicolumn{3}{c|}{Tanks\&Temples} & \multicolumn{3}{c}{Deep Blending}  \\
    
    Method \& Metrics& PSNR $\uparrow$ & SSIM$\uparrow$ & LPIPS$\downarrow$ & PSNR$\uparrow$ & SSIM$\uparrow$ & LPIPS$\downarrow$ & PSNR$\uparrow$ & SSIM$\uparrow$ & LPIPS$\downarrow$ \\
      \midrule   \midrule 
    Instant-NGP~\cite{instant-ngp} & 26.43 & 0.725 & 0.339 & 21.72 & 0.723 & 0.330 & 23.62 & 0.797 & 0.423 \\
    Plenoxels~\cite{fridovich2022plenoxels} & 23.62 & 0.670 & 0.443 & 21.08 & 0.719 & 0.379 & 23.06 & 0.795 & 0.510 \\

    Mip-NeRF 360~\cite{mip-nerf360}&29.23&0.844 &0.207 &22.22 &0.759 &0.257&29.40 & 0.901&0.245\\
    
    2DGS\cite{huang20242d}& 28.98 &0.867 &0.185 & 23.43&0.845&0.181& 29.70 &0.902&0.250\\
    
    Fre-GS~\cite{zhang2024fregs}& 27.85 &0.826 & 0.209 &23.96 &0.841 &0.183  & \cellcolor{yellow!30}29.93 &0.904 & 0.240 \\

    GES~\cite{hamdi2024ges}&28.69 &0.857& 0.206&23.35&0.836&0.198&29.68&0.901&0.252\\

            SeeLe~\cite{huang2025seele} &27.72& 0.814&0.216 &24.02 &0.851 &0.167 &29.79 &0.903 & 0.240\\

    \midrule

  3DGS~\cite{kerbl2023gaussiansplatting} &28.69 &0.870&0.182&23.14 & 0.841&0.183 &29.41&0.903 & 0.243\\
  
   \cellcolor{gray!30} 3DGS (\textbf{+Ours}) & \cellcolor{yellow!30}29.61    & \cellcolor{yellow!30}0.873 
  & \cellcolor{yellow!30}0.173  &\cellcolor{yellow!30}{24.44} & \cellcolor{yellow!30}0.865 & \cellcolor{yellow!30}0.143& 29.74 &  \cellcolor{orange!30}0.909 &\cellcolor{orange!30}0.221  \\

        Scaffold-GS~\cite{lu2024scaffold} &28.84& 0.848&0.220 &23.96 &0.853 & 0.177 &\cellcolor{orange!30}30.21 &\cellcolor{yellow!30}0.906 & 0.254\\

     \cellcolor{gray!30} Scaffold-GS(\textbf{+Ours}) & \cellcolor{orange!30}29.82 & \cellcolor{orange!30}0.877 & \cellcolor{orange!30}0.171& \cellcolor{orange!30}25.54 & \cellcolor{orange!30}0.902 & \cellcolor{orange!30}0.093 & \cellcolor{red!30}30.37 &  \cellcolor{red!30}0.915 &  \cellcolor{red!30}0.153 \\

       3D-HGS~\cite{3dhgs} &29.66& 0.873&0.178 &24.45 & 0.857 & 0.169 & 29.76 &0.905 & 0.242\\

     \cellcolor{gray!30} 3D-HGS(\textbf{+Ours}) & \cellcolor{red!30}30.21 & \cellcolor{red!30}0.878 & \cellcolor{red!30}0.162& \cellcolor{red!30}25.56 & \cellcolor{red!30}0.903 & \cellcolor{red!30}0.091 & 29.74 &  \cellcolor{yellow!30}0.906 &  \cellcolor{yellow!30}0.236 \\

    \bottomrule
    \end{tabular}
}

\label{t1_rec}

\end{table}

\begin{table}[t]
\caption{\textbf{Comparisons of the number of 3D Gaussians and training time.} Num. denotes the number of 3D Gaussian parameters. We evaluate these results on the Mip-NeRF 360 dataset. }
\centering
\scriptsize
\def\arraystretch{1.2}
\resizebox{0.95\linewidth}{!}{
\begin{tabular}{l|cc|cc|cc}
\toprule
Method        & 3DGS & 3DGS + Ours & Scaffold & Scaffold + Ours & 3D-HGS & 3D-HGS + Ours \\ \midrule \midrule
Num. $\times10^6$ & 1.1  & \textbf{0.8}        & 0.4         & \textbf{0.3}     & 0.7   & \textbf{0.5}      \\ 
 Time (h)& \textbf{0.3}& 2.4& \textbf{0.2}& 1.5& \textbf{0.4}&2.6\\    
 PSNR          &   28.69   &    \textbf{ 29.61}        &      28.84       &     \textbf{29.82 }    &     29.66  &     \textbf{30.21  }          \\   \bottomrule

\end{tabular}
}
\label{t_parameter}

\end{table}

\section{Experiments}

\subsection{General Object Novel View Synthesis}
To assess the performance of our proposed approach, we compare our improved version on 3DGS~\cite{kerbl2023gaussiansplatting}, Scaffold-GS~\cite{lu2024scaffold}, and 3D-HGS~\cite{3dhgs} baselines with their original methods. 
We also compare our method with recent state-of-the-art 3D NVS methods, including Instant-NGP~\cite{instant-ngp}, Plenoxels~\cite{fridovich2022plenoxels}, Mip-NeRF 360~\cite{mip-nerf360}, 2DGS~\cite{2dgs}, Fre-GS~\cite{zhang2024fregs}, GES~\cite{hamdi2024ges}, and SeeLe~\cite{huang2025seele}. 
The quantitative results are shown in Table \ref{t1_rec}.
We conduct general object NVS experiments on three commonly used datasets, such as Mip-NeRF 360~\cite{mip-nerf360}, Tank\&Temples~\cite{tanktemple}, and Deep Blending~\cite{deepblending}.
In Table \ref{t1_rec}, it can be observed that our method, integrated into 3DGS, Scaffold-GS, and 3D-HGS, achieves state-of-the-art results in terms of PSNR, SSIM, and LPIPS. 
In particular, Tank\&Temples~\cite{tanktemple} is a more challenging dataset than the others, containing more challenging scenes with the presence of texture-less regions, lighting changes, and reflections.

\begin{figure*}[t!]

    \centering
    \includegraphics[width=0.99\linewidth]{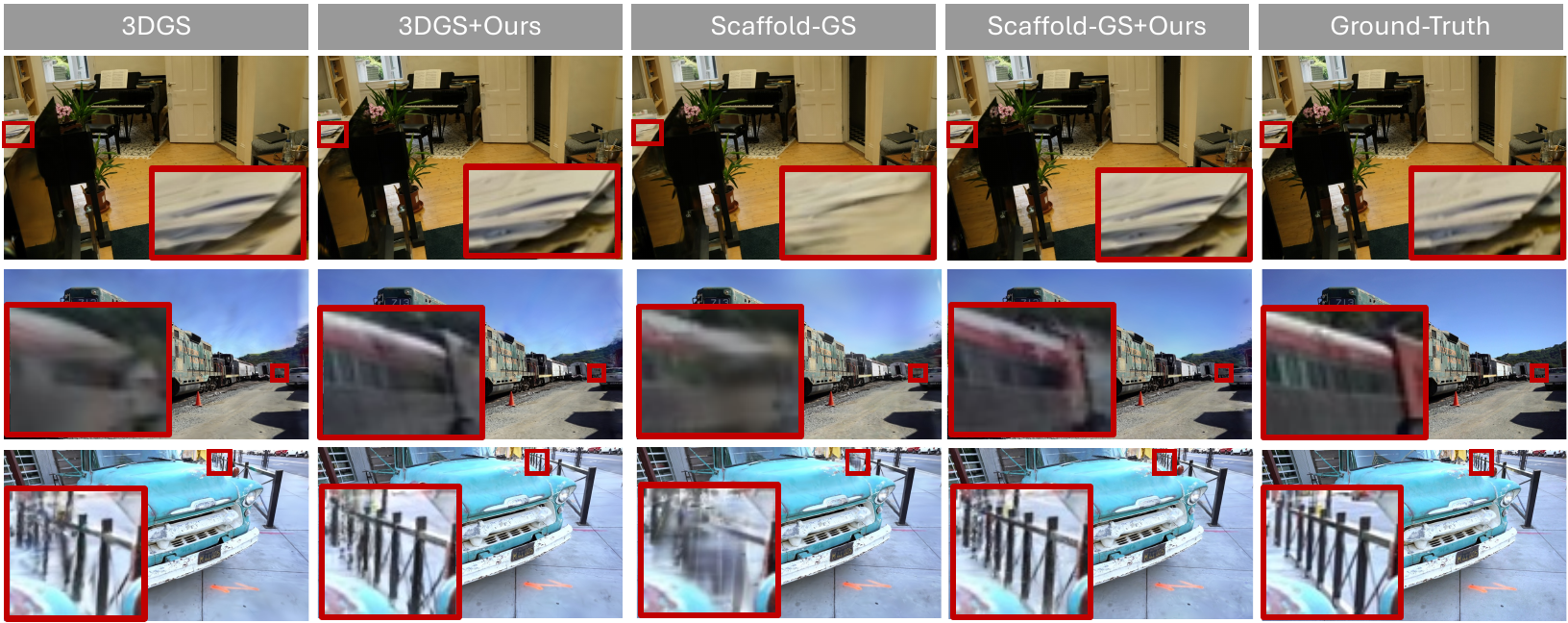} 
    \caption{ \textbf{Qualitative comparisons of 3DGS~\cite{kerbl2023gaussiansplatting}, Scaffold-GS~\cite{lu2024scaffold}, and their improved version integrating our method across various datasets.} We use \textcolor{red}{red} close-up patches to highlight the visual differences for clearer visibility. We can observe that our proposed method improves the original 3DGS and Scaffold-GS for challenging scenes with drastically varied lighting effects, strong reflection, and fine details. }
    \label{vis_rec}
    
\end{figure*}

As for qualitative comparisons, we provide the results in Fig.~\ref{vis_rec}, through comparisons of 3DGS, Scaffold-GS, and their methods integrated with our method. 
It can be observed that our method can improve the novel view synthesis performance quantitatively and qualitatively.  In particular, previous methods are struggling to deal with scenes with strong reflection, fine details, and strong lighting changes, leading to the phenomena of floaters, distortion, and over-smoothness. 
In contrast, our proposed multi-view regulated learning imposes multi-view constraints on the learning phase of 3D Gaussians to render novel views more accurately.

\begin{figure*}[t!]
    \centering
    \includegraphics[width=0.99\linewidth]{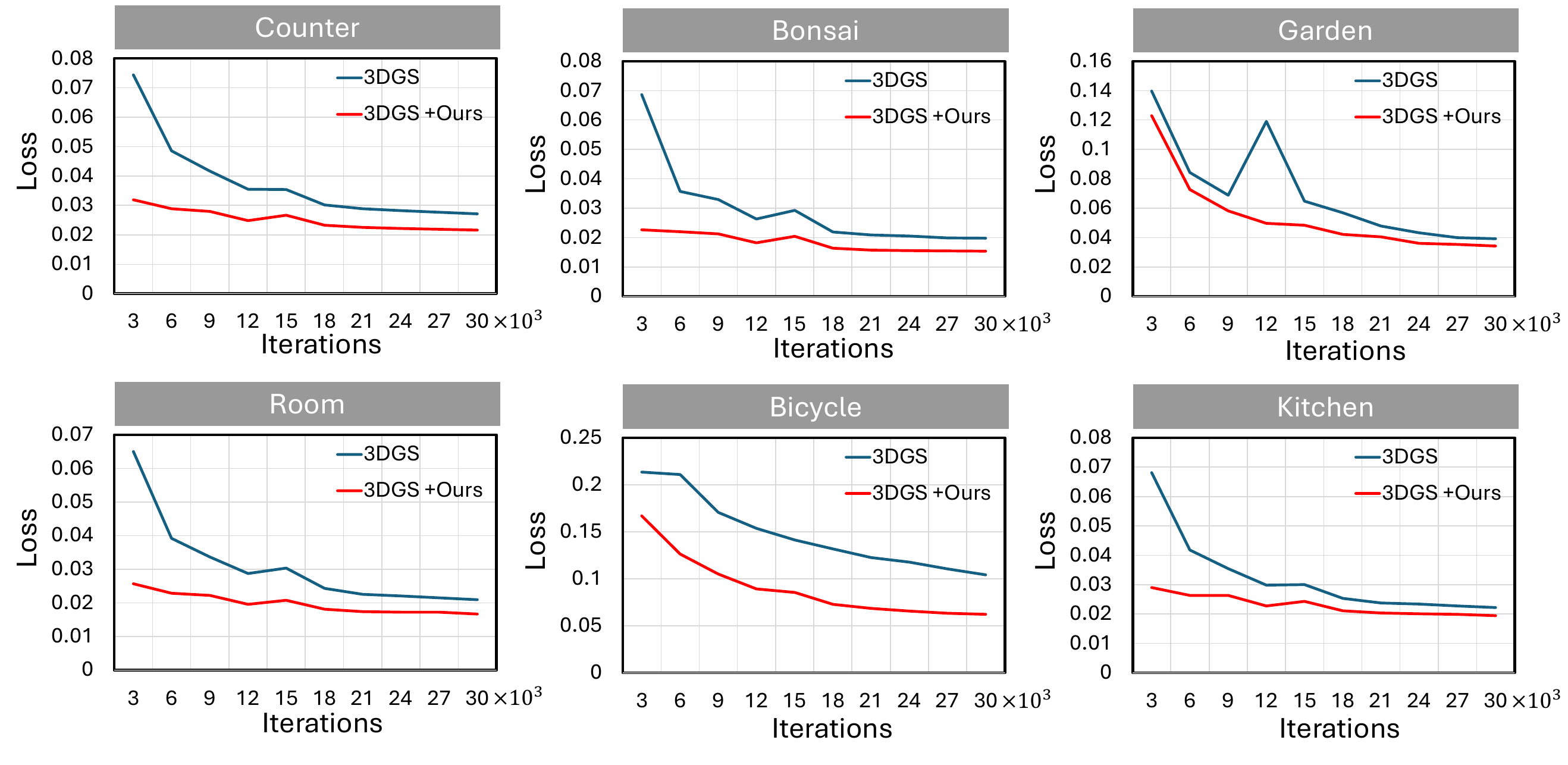} 
    \caption{\textbf{Training loss per scene visualizations of 3DGS and our improved version of 3DGS on the Mip-NeRF 360 dataset~\cite{mip-nerf360}.}  We demonstrate that our proposed method facilitates the optimization of 3DGS and achieves better results. }
    \label{fig:loss}
    
\end{figure*}

As shown in Table~\ref{t_parameter}, MVGS integrated into 3DGS, Scaffold-GS, and 3D-HGS can achieve better results with fewer Gaussian parameters. This is attributed to the proposed multi-view training strategy that not only imposes multi-view constraints but also makes the structure more compact. Although it leads to more training costs, it can achieve higher-quality NVS results and obtain better trade-offs between training time, rendering quality, and the Gaussian number.
These results indicate that previous methods integrated with our method can achieve better qualitative results and reconstruct more satisfactory details. Moreover, we also demonstrate that our method accelerates the convergence and stabilizes the optimization process, as shown in Fig.~\ref{fig:loss}. Compared to 3DGS, our method achieves better performance with the stable gradient descent process. These results demonstrate the effectiveness of our proposed method.

\subsection{Ablation Study}

\begin{figure*}[t]
    \centering
    \includegraphics[width=0.99\linewidth]{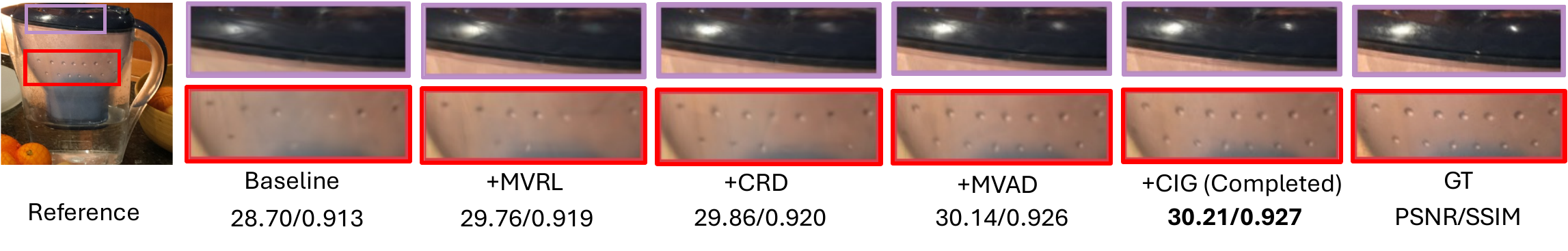} 
    \caption{\textbf{Visual comparisons of the progressive ablation studies for the proposed components. } We employ 3DGS as our baseline and improve it by progressively integrating our proposed components. It can be observed that the novel view synthesis performance is gradually improved.}
    \label{vis_ablation}
   
\end{figure*}

To comprehensively demonstrate the effectiveness of our proposed method, we conduct extensive ablation studies to evaluate the contribution of each component. 
As outlined in our method section, our proposed \emph{MVGS} consists of four key components, including multi-view regulated learning, cross-ray densification,  multi-view augmented densification, and cross-intrinsic guidance.
We present a detailed ablation study in Fig. \ref{vis_ablation}.
We leverage 3DGS as our baseline and integrate our proposed components progressively into it to demonstrate the effectiveness of the proposed methods. 
Specifically, incorporating the proposed multi-view regulated learning (MVRL) into 3DGS imposes the multi-view constraint for the model to learn in a more stable and accurate way.
After that, we also progressively embed our proposed cross-ray densification (CRD) method into the baseline, enforcing 3D Gaussians to fit complex structures for better results. 
When the multi-view augmented densification (MVAD) is employed, Gaussians are enhanced to facilitate multi-view training. As we can see, performance improves significantly.
Finally, when we adopt our proposed cross-intrinsic guidance (CIG) strategy, the model captures finer details through multi-scale training, leading to higher rendering quality.
These results demonstrate the effectiveness of our components.

To further demonstrate the general applicability of the proposed components, we conduct detailed ablation studies across various Gaussian-based 3D Novel View Synthesis methods as shown in Table~\ref{t_ablation}. To be specific, we utilize three representative methods, including 3DGS~\cite{kerbl2023gaussiansplatting}, Scaffold-GS~\cite{lu2024scaffold}, and 3D-HGS~\cite{3dhgs} as baselines and integrate our proposed method into them. As depicted in Table~\ref{t_ablation}, the original performance of these baselines is inferior. When we incorporate the proposed multi-view regulated learning (MVRL) into baselines, the performance is largely improved.  This significant improvement is due to the proposed MVRL imposing multi-view constraints on the optimization of 3D Gaussians to enhance 3D Gaussian multi-view consistency.
In addition, we also propose two novel densification strategies,  cross-ray densification and multi-view augmented densification, to enhance 3D Gaussian primitives in specific regions for better fitting with the multi-view supervision. To fully leverage multi-view information, we propose cross-intrinsic guidance to train models with an image pyramid for accommodating more views for multi-view training. 
When all components are combined, these Gaussian-based methods achieve state-of-the-art results for high-quality novel view synthesis. 
These results further demonstrate the effectiveness of our proposed method and also indicate that our method can consistently enhance the existing Gaussian-based methods to reach state-of-the-art performance.

\begin{table}[t!]
\centering
\caption{\textbf{Detailed ablation studies across various Gaussian-based methods. } We present ablation studies on state-of-the-art 3D reconstruction methods improved by our proposed method, including 3DGS~\cite{kerbl2023gaussiansplatting}, Scaffold-GS~\cite{lu2024scaffold}, and 3D-HGS~\cite{3dhgs}.   We report results on the Mip-NeRF 360 dataset~\cite{mip-nerf360}.}
\scriptsize
\def\arraystretch{1.2}
\resizebox{0.95\linewidth}{!}
{
\begin{tabular}{l|ccccccccc}
\toprule \toprule
\multirow{2}{*}{Method}             & \multicolumn{3}{c|}{3DGS}                  & \multicolumn{3}{c|}{Scaffold-GS} & \multicolumn{3}{c}{3D-HGS} \\ \cline{2-10} 
                                    & PSNR  & SSIM & \multicolumn{1}{c|}{LPIPS} &   PSNR  & SSIM & \multicolumn{1}{c|}{LPIPS}  &PSNR  & SSIM & LPIPS     \\ \hline
Baseline                            & 28.69&  0.870& \multicolumn{1}{c|}{0.182}      &  28.84&  0.848& \multicolumn{1}{c|}{0.220}  &     29.66&   0.873&  0.178\\
+Multi-view regulated learning         & 29.26&  \cellcolor{yellow!30}0.871& \multicolumn{1}{c|}{0.179}      &  29.47&  0.861& \multicolumn{1}{c|}{0.189}  &      29.94&    0.875&  0.174\\
+Cross-ray densification            & \cellcolor{yellow!30}29.37&   \cellcolor{orange!30}0.872& \multicolumn{1}{c|}{\cellcolor{yellow!30}0.178}      & \cellcolor{yellow!30}29.53&  \cellcolor{yellow!30}0.863& \multicolumn{1}{c|}{\cellcolor{yellow!30}0.184}  &     \cellcolor{yellow!30}30.01&     \cellcolor{yellow!30}0.876&  \cellcolor{yellow!30}0.171\\
+Multi-view augmented densification & \cellcolor{orange!30}29.52&   \cellcolor{orange!30}0.872& \multicolumn{1}{c|}{\cellcolor{orange!30}0.175}      &  \cellcolor{orange!30}29.73& \cellcolor{orange!30}0.869& \multicolumn{1}{c|}{\cellcolor{orange!30}0.176}  &     \cellcolor{orange!30}30.15&   \cellcolor{orange!30}0.877&   \cellcolor{orange!30}0.165\\
+Cross-intrinsic guidance (full)      & \cellcolor{red!30}29.61&   \cellcolor{red!30}0.873& \multicolumn{1}{c|}{\cellcolor{red!30}0.173}      &  \cellcolor{red!30}29.82& \cellcolor{red!30}0.877& \multicolumn{1}{c|}{\cellcolor{red!30}0.171}  &     \cellcolor{red!30}30.21&        \cellcolor{red!30}0.878&   \cellcolor{red!30}0.162\\ \bottomrule
\end{tabular}}
\label{t_ablation}

\end{table}

\begin{figure*}[t]
    \centering
    \includegraphics[width=0.99\linewidth]{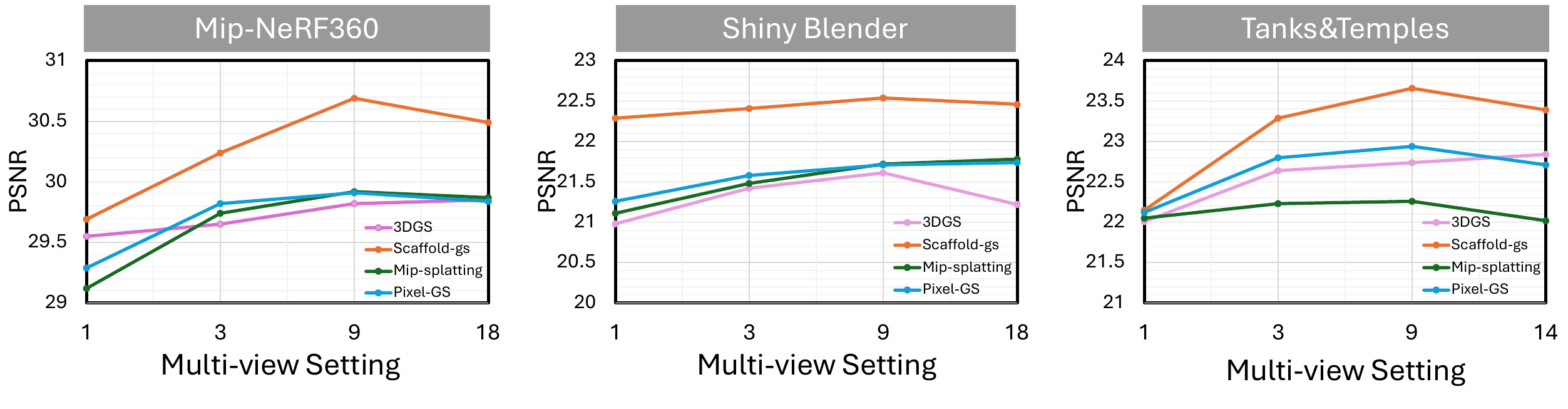} 
    \caption{\textbf{Analysis of the multi-view training settings.} 
    We improve four representative state-of-the-art Gaussian-based methods with the proposed multi-view regulated training. We report results on three representative datasets. When the number of views is one, it means the original single-view training manner, while our proposed multi-view training achieves better results than previous single-view training.
    }
    \label{vis_mv}
  
\end{figure*}

\subsection{Analysis on Multi-view Training Settings}
In our experiments, we found that the appropriate multi-view training configurations significantly improve rendering performance compared to existing Gaussian-based methods. 
This improvement is a notable characteristic of our proposed method.
As shown in Fig. \ref{vis_mv}, we compare existing state-of-the-art Gaussian methods with their counterparts enhanced by our proposed multi-view training.
Fig. \ref{vis_mv} investigates the relation between rendering improvement and the number of views in multi-view training.
We observe that incorporating our multi-view training into existing methods leads to a substantial improvement in novel view synthesis quality. 
This enhancement is primarily attributed to our proposed multi-view regulated learning that constrains the optimization of the entire 3D Gaussians with multi-view information, thus enforcing global consistency across views. 
However, when the number of multi-views increases to a certain number, the performance begins to degrade.
This occurs because an excessive number of multi-views leads to a large number of sampled views analogous to a region of views, encouraging 3D Gaussians to overfit in an area of the scene. 
Therefore, a moderate or scanty multi-view setting is more conducive to the optimization.

\section{Conclusion}
In this work, we propose a novel and universal optimization method, dubbed MVGS, to improve the Novel View Synthesis performance for existing Gaussian-based methods. The core of MVGS lies in the proposed multi-view regulated learning, constraining the optimization of 3D Gaussians with multi-view information.  We show that our method can be integrated into existing methods to achieve state-of-the-art rendering performance. We further demonstrate that our proposed cross-intrinsic guidance scheme and multi-view cross-ray densification facilitate multi-view training for better results. 
We believe the plug-and-play nature of MVGS opens up new avenues for accelerating high-fidelity Gaussian Splatting optimization in complex and real-world environments.




%
%

\section*{Acknowledgements}
This research is funded in part by ARC-Discovery grant (DP220100800 to XY), ARC-DECRA grant (DE230100477 to XY), NVIDIA academic grant program, and Google Research Scholar Program. We thank all anonymous reviewers and ACs for their constructive suggestions.

\appendix

 \section{Discussion and Limitations}

Extensive experiments demonstrate that our method is a plug-and-play optimizer, which can be readily integrated with existing Gaussian-based methods, making it a versatile solution for photorealistic rendering.
Although our approach significantly improves the quantitative and qualitative results of existing methods across various datasets, including challenging scenes with dynamic lighting and reflections, it also incurs longer training time and higher computational cost.
The incorporation of multi-view constraints entails optimizing across multiple views per iteration, leading to higher computational cost compared to single-view optimization paradigms. 
Moreover, our densification strategies, while effective in enhancing detail, may introduce additional complexity in cases where simpler scenes are involved.

\noindent\textbf{Future Work:}
To address the training time issue, future work will explore more efficient optimization techniques, such as adaptive multi-view selection and progressive multi-view training, to dynamically balance computational cost and performance. Additionally, developing a more efficient GPU parallel implementation may further improve computational efficiency.
Moreover, investigating hybrid approaches that combine explicit Gaussian representations with implicit neural field modeling could offer a balance between training efficiency and rendering quality.

\section{Additional Details of Multi-view Cross-ray Densification}

As shown in Algorithm~\ref{alg:cross_ray_densification}, the proposed cross-ray densification enhances the multi-view constraint by leveraging the ray tracing technique. 
In summary, the proposed multi-view cross-ray densification lies in the following strategies: 
(1) Adaptive Density: When the threshold $\hat{\beta}$ is the same as the original one, it prevents over-densification in simple regions.  When the threshold $\hat{\beta}$ is half of the original threshold, it enhances 3D Gaussians fitting toward complex areas;
(2) Multi-View Enhancement: The ray casting technology from multiple views allows us to locate the 3D regions containing a set of 3D Gaussians that contribute significantly to these views. In this way, we can enhance the densification effect of these 3D Gaussians for multi-view enhancement;
(3) Photometric Guidance: We choose the loss to locate the lowest-quality regions since it directly highlights the texture-dense regions that should be improved.

\begin{algorithm}[h]
 \textbf{Input: }  Multi-view loss maps $\mathcal{L}$, Multiple view setting $M$, sliding window size $(h, w)$, and number of rays $n_r = 4$ per window.  \\
  \textbf{Output: }  Densified 3D Gaussians.  \\
\SetKwFor{ForEach}{for each}{do}{end}
\textbf{Step 1: Loss Map Analysis} \\
\ForEach{view $m \in [1, M]$}{
    Calculate the loss map $\mathcal{L}_m$. 
}
\textbf{Step 2: Region Localization}\\
\ForEach{loss map $\mathcal{L}_m$}{
    Compute the average loss within a sliding window.\\
    Identify regions $\mathcal{R}$ with the largest average loss values.
}
\textbf{Step 3: Ray Casting and Intersection}\\
\ForEach{region $r \in \mathcal{R}$}{
    Cast $n_r$ rays per perspective from the vertices of $r$. \\
    Compute intersected planes of rays across different views.\\
    Form cuboids from intersections.
}
\textbf{Step 4: Densification}\\
\ForEach{cuboid } {
    Identify 3D Gaussians within each cuboid.\\
    Densify these 3D Gaussians.
}
\textbf{Return:} Densified 3D Gaussians.
\caption{Cross-ray Densification}
\label{alg:cross_ray_densification}

\end{algorithm}

\section{Implementation}

We conduct extensive experiments on various tasks to demonstrate that our method can improve the performance of each baseline approach, ranging from static synthetic object-level scenes to indoor, outdoor, large-scale, and dynamic scenes. 
Evaluation results on each dataset prove that our method outperforms the baselines,  especially in challenging cases, such as insufficient observations, texture-less areas, view-dependent lighting effects, and fine-scale details.

\subsection{Dataset and Metrics}
We conduct comprehensive comparisons across 31 scenes from public datasets. 
To demonstrate the effectiveness of our method, we evaluate our MVGS on several popular tasks, such as 3D Novel View Synthesis (NVS), reflective object NVS, and 4D NVS.
For 3D NVS, we evaluate our MVGS on several widely used scenes following Scaffold-GS~\cite{lu2024scaffold}, including seven scenes from Mip-NeRF360~\cite{mip-nerf360}, two scenes from Deep Blending~\cite{deepblending}, and two scenes from Tank\&Temples~\cite{tanktemple}.
For the task of reflective object Novel View Synthesis, we evaluate our method on Shiny Blender~\cite{refnerf} and Glossy Synthetic~\cite{liu2023nero} datasets. 
As for 4D Novel View Synthesis, the D-NeRF dataset~\cite{dnerf} is employed for evaluation. In addition to the widely used metrics, such as PSNR, SSIM, and LPIPS, we additionally report the rendering speed (FPS) and storage size (MB) for rendering efficiency and model compactness.

\subsection{Baseline and Implementation}
In our experiments, we utilize novel view synthesis metrics like PSNR, SSIM, and LPIPS to evaluate the performance of models.
For general object Novel View Synthesis, 3DGS~\cite{kerbl2023gaussiansplatting} and Scaffold-GS~\cite{lu2024scaffold} are selected as our baselines due to their state-of-the-art performance.
For reflective object Novel View Synthesis, we choose 3DGS-DR~\cite{3dgsdr} as our main baseline since it is the recent state-of-the-art method to reconstruct glossy objects.
As for 4D Novel View Synthesis, 4DGS~\cite{4dgs} is selected as our baseline due to its fast rendering speed and high-quality 4D Novel View Synthesis performance.
In large-scale scene Novel View Synthesis, Octree-GS~\cite{ren2024octree} is adopted as our baseline since its level-of-detail structure is suitable for large-scale scenes.
In our proposed method, we set $M_s$ = $\{48, 24,12,8 \}$ and $\tau = 1$. 
As for the other setting, we follow the implementation setting of these baselines.  
We conduct our experiments on the RTX 3090 Ti GPU.

\section{Additional Experiments}

\subsection{Training with More Iterations}
\begin{figure*}[t]
    \centering
    \includegraphics[width=0.95 \linewidth]{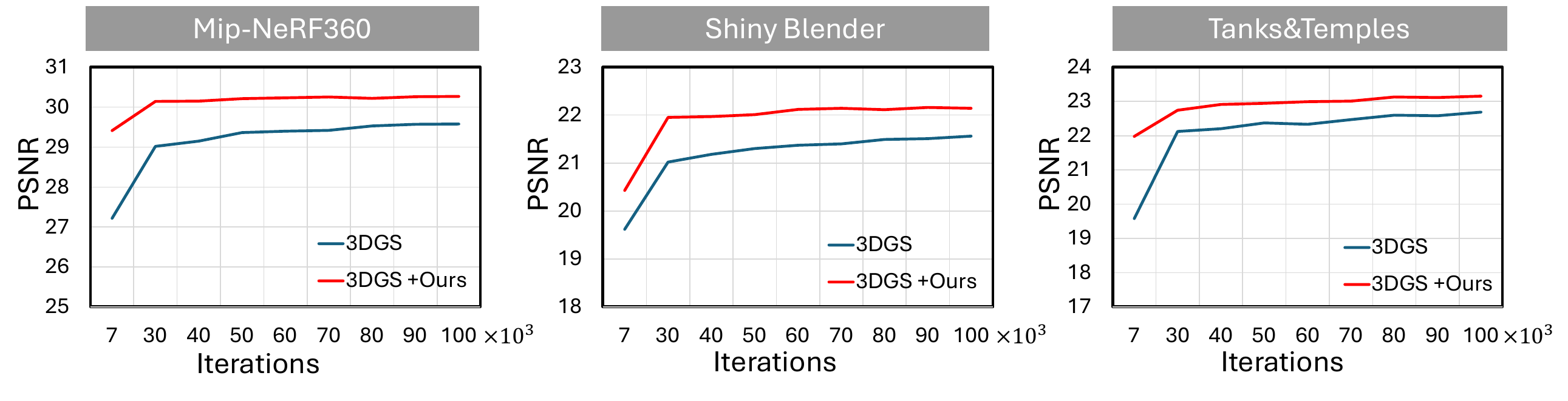} 
    \caption{\textbf{Study on the effect of training with more time.}
    We leverage state-of-the-art 3DGS~\cite{kerbl2023gaussiansplatting} as our baseline and conduct experiments on three representative datasets, such as Mip-NeRF 360~\cite{mip-nerf360}, Shiny Blender~\cite{refnerf}, and Tanks\&Temples~\cite{tanktemple}.
    }
    \label{vis_iters}
        
\end{figure*}

For a fair comparison, we also evaluate the impact of training with more iterations.  As illustrated in Fig.~\ref{vis_iters}, we conduct experiments on three representative datasets, such as Mip-NeRF 360~\cite{mip-nerf360}, Shiny Blender~\cite{refnerf}, and Tanks\&Temples~\cite{tanktemple} with 3DGS~\cite{kerbl2023gaussiansplatting} as our baseline and its improved version by integrating with our proposed method. As displayed in Fig.~\ref{vis_iters}, the performance of 3DGS is obviously inferior to ours. 
Despite being trained for more iterations, 3DGS consistently fails to outperform our method. It indicates that training with more time does not compensate for the multi-view constraint and structural enhancements absent in the original 3DGS.
In contrast, our method not only speeds up the training convergence but also delivers superior results. 
These results demonstrate the significance of our proposed method and indicate that the original 3DGS training with more time cannot achieve better performance than ours.

\subsection{Analysis on Sum and Average Operations}
\begin{figure*}[t]
    \centering
    \includegraphics[width=0.95\linewidth]{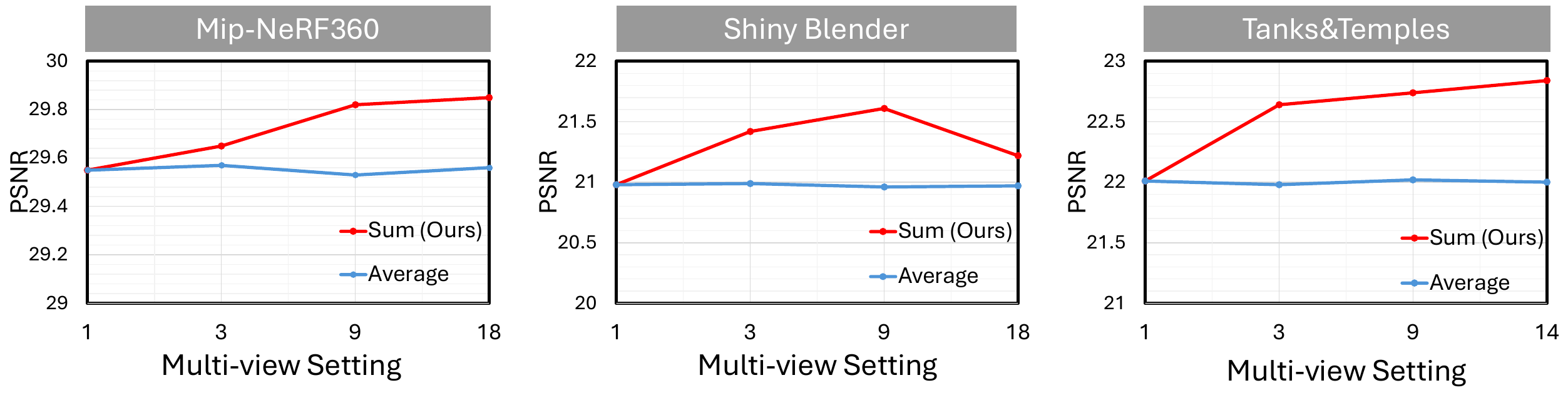} 
    \caption{\textbf{Analysis on the Sum and Average operation in multi-view learning.} We employ 3DGS to evaluate whether the multi-view loss should use the Sum or Average operation on three commonly used datasets. 
    }
    \label{vis_sum_avg}
\end{figure*}

In our proposed multi-view training, we use the Sum operation to aggregate gradients, not Average. Here, we conduct comparisons between the Sum and Average operations with 3DGS. As shown in Fig.~\ref{vis_sum_avg}, when the number of multi-view training settings increases, the Sum operation improves rendering results largely compared with the Average operation. This improvement arises because the Sum operation preserves and amplifies the gradient magnitude, whereas the Average operation reduces it. Consequently, a larger gradient magnitude enhances the optimization process for more powerful multi-view coherence constraints in multi-view training, leading to superior Novel View Synthesis results.

\subsection{Additional Results on 3D Reconstruction}
\begin{figure*}[t]
    \centering
    \includegraphics[width=1 \linewidth]{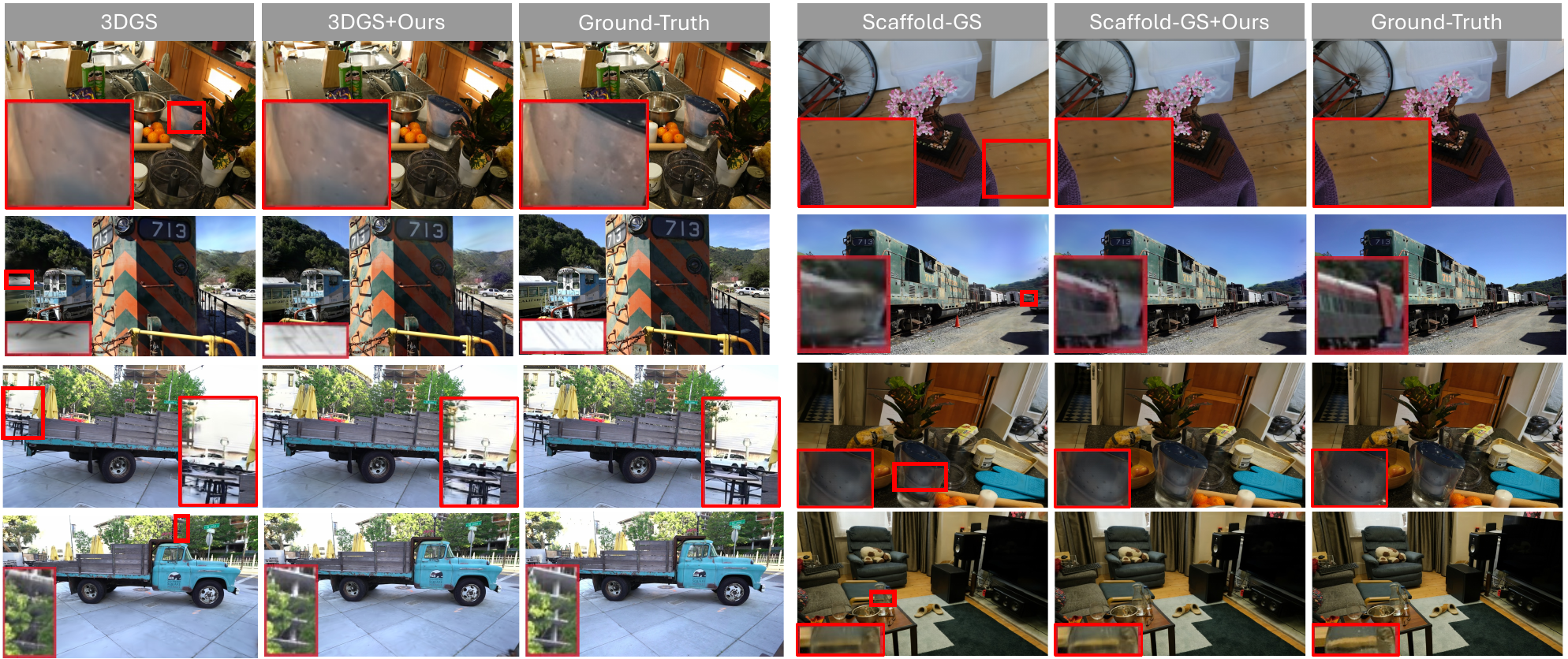} 
    \caption{\textbf{Additional qualitative comparisons of general object reconstruction. We compare 3DGS~\cite{kerbl2023gaussiansplatting} and Scaffold-GS~\cite{lu2024scaffold} with their improved version by integrating our method across various datasets.} We employ \textcolor{red}{red} close-up patches to highlight the visual differences for better differentiation. It can be observed that our proposed method can improve the original 3DGS and Scaffold-GS for challenging scenes. }
    \label{vis_rec_supp}
\end{figure*}

In this section, we present additional experimental results for 3D reconstruction. To sufficiently demonstrate the effectiveness of our proposed method, we showcase per-scene quantitative results of the Mip-NeRF 360 dataset~\cite{mip-nerf360} in Table \ref{t1_supp}. As we can see in Table \ref{t1_supp}, 3DGS~\cite{kerbl2023gaussiansplatting} and Scaffold-GS~\cite{lu2024scaffold} integrated with our proposed method are better than their original performance. It demonstrates the effectiveness of our proposed method to improve 3D reconstruction results. We also present per-scene results of Tank\&Temples~\cite{tanktemple} and Deep Blending~\cite{deepblending} in Table \ref{t2_supp}. To be specific, we select representative scenes, including Truck and Train from Tank\&Temples, and Playroom and Drjohnson from Deep Blending, respectively. It can be observed that our proposed method also demonstrates superior performance. In addition, we display additional visual comparisons of the task of 3D reconstruction in Fig. \ref{vis_rec_supp}. We observe the original 3DGS and Scaffold-GS cannot recover details of the transparent surface or far objects. By integrating our proposed method, our proposed multi-view constraint encourages 3D Gaussians to capture finer details of multiple views and improve reconstruction quality.

\begin{table*}[t]
\centering
\caption{\textbf{Detailed quantitative results of state-of-the-art 3D reconstruction methods on Mip-NeRF 360 dataset~\cite{mip-nerf360}}. The \colorbox{red!30}{best}, \colorbox{orange!30}{second best}, and \colorbox{yellow!30}{third best} results are denoted by red, orange, and yellow, respectively. }
\scriptsize
\def\arraystretch{1.2}
\begin{adjustbox}{width=\linewidth}
    \begin{tabular}{ l | l l l | l l l| l l l| l l l }
    \toprule

    Metrics& PSNR $\uparrow$ & SSIM$\uparrow$ & LPIPS$\downarrow$ & PSNR$\uparrow$ & SSIM$\uparrow$ & LPIPS$\downarrow$ & PSNR$\uparrow$ & SSIM$\uparrow$ & LPIPS$\downarrow$ & PSNR$\uparrow$ & SSIM$\uparrow$ & LPIPS$\downarrow$ \\
  3D Scenes & \multicolumn{3}{c|}{Stump} & \multicolumn{3}{c|}{Room} & \multicolumn{3}{c|}{Counter} & \multicolumn{3}{c}{bonsai} \\
    \midrule \midrule
        Mip-NeRF 360~\cite{mip-nerf360}&\cellcolor{yellow!30}26.40 &0.744 &0.261 & 31.63& 0.913& {0.211}& \cellcolor{yellow!30}{29.55}& 0.894& 0.204 & \cellcolor{orange!30}{33.46}& 0.941 &\cellcolor{yellow!30}0.176\\
        
    3DGS~\cite{kerbl2023gaussiansplatting} &\cellcolor{orange!30}{26.55}  &\cellcolor{yellow!30}{0.775} &\cellcolor{red!30}{0.210} &  30.63& 0.914& 0.220&  28.70&  0.905&0.204&  31.98& 0.938& 0.205\\

    Scaffold-GS~\cite{lu2024scaffold} &26.27 &\cellcolor{red!30}{0.784} &0.284 & \cellcolor{yellow!30}{31.93}& \cellcolor{yellow!30}{0.925}& \cellcolor{yellow!30}{0.202}& 29.34& \cellcolor{yellow!30}{0.914}& \cellcolor{yellow!30}{0.191}& 32.70& \cellcolor{yellow!30}0.946& 0.185\\
    \cline{1-13}

     3DGS (\textbf{+Ours}) &{26.39} &\cellcolor{orange!30}0.760& \cellcolor{yellow!30}{0.243}& \cellcolor{orange!30}{32.84}& \cellcolor{orange!30}{0.932}& \cellcolor{orange!30}0.184&  \cellcolor{orange!30}{30.21}&  \cellcolor{orange!30}{0.928}&  \cellcolor{orange!30}{0.151}& \cellcolor{yellow!30}33.05& \cellcolor{orange!30}{0.949}& \cellcolor{orange!30}{0.167}\\

    Scaffold-GS(\textbf{+Ours})  & \cellcolor{red!30}26.74 &\cellcolor{yellow!30} 0.775 & \cellcolor{orange!30}0.232& \cellcolor{red!30}33.08 &\cellcolor{red!30}0.935 & \cellcolor{red!30}0.174 &\cellcolor{red!30}30.98 & \cellcolor{red!30}0.929&\cellcolor{red!30}0.149& \cellcolor{red!30}33.69& \cellcolor{red!30}0.953&\cellcolor{red!30}0.163 \\
    \midrule \midrule
   
    3D Scenes & \multicolumn{3}{c|}{Bicycle} & \multicolumn{3}{c|}{Garden} & \multicolumn{3}{c|}{Kitchen}  \\
    

    \cline{1-10}
    Mip-NeRF 360~\cite{mip-nerf360}&24.37&0.685&\cellcolor{yellow!30}0.301&26.98&0.813&0.170&\cellcolor{orange!30}{32.23}&0.920&0.127\\
        
    3DGS \cite{kerbl2023gaussiansplatting} & \cellcolor{red!30}{25.25} &\cellcolor{red!30}{0.771} &\cellcolor{red!30}{0.205} & \cellcolor{orange!30}{27.41} & \cellcolor{red!30}{0.868} & \cellcolor{red!30}{0.103} & 30.32 & 0.922 & 0.129\\

    Scaffold-GS~\cite{lu2024scaffold} &24.50&0.705&0.306&27.17&0.842&0.146&31.30&\cellcolor{yellow!30}{0.928}&\cellcolor{yellow!30}{0.126}\\
    \cline{1-10}

     3DGS (\textbf{+Ours}) &\cellcolor{yellow!30}{25.08} &\cellcolor{yellow!30}{0.752}& \cellcolor{orange!30}{0.226}&\cellcolor{yellow!30}{27.23}&\cellcolor{orange!30}{0.856}& \cellcolor{orange!30}{0.123}& \cellcolor{red!30}{32.57}& \cellcolor{red!30}{0.934}& \cellcolor{red!30}{0.113}\\

    Scaffold-GS(\textbf{+Ours})& \cellcolor{orange!30}25.23 &\cellcolor{orange!30}0.760 &\cellcolor{orange!30}0.226 & \cellcolor{red!30}27.48 & \cellcolor{yellow!30}0.855 &\cellcolor{yellow!30}0.124 &\cellcolor{yellow!30}31.96 &\cellcolor{orange!30}0.933 & \cellcolor{orange!30}0.114 \\

    \end{tabular}
\end{adjustbox}

\label{t1_supp}
\end{table*}

\begin{table*}[t]
\caption{\textbf{Detailed quantitative comparisons of state-of-the-art 3D reconstruction methods on Tank\&Temples~\cite{tanktemple} and Deep Blending~\cite{deepblending}.
} We choose two challenging scenes, Truck and Tran from the Tank\&Temples dataset for evaluation. As for Deep Blending, we select two representative scenes, Playroom and Drjohnson for assessment.
}

\scriptsize
\def\arraystretch{1.2}
\begin{adjustbox}{width=1\linewidth}
    \begin{tabular}{ l | l l l | l l l| l l l|l l l }
     \toprule
    \multirow{ 2}{*}{}Dataset& \multicolumn{6}{c|}{Tanks\&Temples} & \multicolumn{6}{c}{Deep Blending} \\ \toprule \toprule
    
    3D Scenes & \multicolumn{3}{c|}{Truck} & \multicolumn{3}{c|}{Train} & \multicolumn{3}{c|}{Playroom} & \multicolumn{3}{c}{Drjohnson}  \\
    
    Method& PSNR $\uparrow$ & SSIM$\uparrow$ & LPIPS$\downarrow$ & PSNR$\uparrow$ & SSIM$\uparrow$ & LPIPS$\downarrow$ & PSNR$\uparrow$ & SSIM$\uparrow$ & LPIPS$\downarrow$ & PSNR$\uparrow$ & SSIM$\uparrow$ & LPIPS$\downarrow$ \\ \midrule 
    3DGS~\cite{kerbl2023gaussiansplatting} & 25.18 &0.879 &0.148 & 21.09 & 0.802 & 0.218 & 30.04 & \cellcolor{yellow!30}{0.906} & \cellcolor{yellow!30}{0.241} & 28.77& 0.899&0.244\\
    Mip-NeRF 360~\cite{mip-nerf360}&24.91&0.857&0.159&19.52&0.660&0.354&29.66&0.900&0.252 &29.14& \cellcolor{yellow!30}0.901& \cellcolor{orange!30}0.237\\

    Scaffold-GS~\cite{lu2024scaffold} &\cellcolor{yellow!30}{25.77}&\cellcolor{yellow!30}{0.883}&\cellcolor{yellow!30}{0.147}&\cellcolor{yellow!30}{22.15}&\cellcolor{yellow!30}{0.822}&\cellcolor{yellow!30}{0.206}&\cellcolor{orange!30}{30.62}&0.904&0.258& \cellcolor{orange!30}29.80& \cellcolor{red!30}0.907& 0.250\\
    \cline{1-13}
    
  3DGS \textbf{(+Ours)} &\cellcolor{orange!30}{26.14} &\cellcolor{orange!30}{0.893}& \cellcolor{orange!30}{0.125}&\cellcolor{orange!30}{22.74}&\cellcolor{orange!30}{0.838}& \cellcolor{orange!30}{0.162}& \cellcolor{yellow!30}30.33&\cellcolor{orange!30}0.927& \cellcolor{orange!30}{0.201} & \cellcolor{yellow!30}29.16& 0.892& \cellcolor{yellow!30}0.241\\
    
    Scaffold-GS\textbf{(+Ours)} & \cellcolor{red!30}27.19 & \cellcolor{red!30}0.926&\cellcolor{red!30}0.071 & \cellcolor{red!30}23.88 & \cellcolor{red!30}0.878&\cellcolor{red!30}0.116 & \cellcolor{red!30}30.84 &\cellcolor{red!30}0.925 &\cellcolor{red!30}0.152 & \cellcolor{red!30}29.91 & \cellcolor{orange!30}0.905 & \cellcolor{red!30}0.154 \\
    \bottomrule

    \end{tabular}
\end{adjustbox}
\label{t2_supp}
\end{table*}

\begin{table*}[t!]

\centering
\scriptsize
\def\arraystretch{1.2}
\caption{\textbf{Quantitative comparisons of state-of-the-art reflective object reconstruction methods.} We demonstrate our method can improve reconstruction performance for challenging reflective scenes.  We report results on Shiny Blender~\cite{refnerf} and Glossy Synthetic datasets~\cite{liu2023nero}. }
\begin{adjustbox}{width=1\linewidth}
\begin{tabular}{cl|cccccc|cccccc}
\toprule
\multicolumn{2}{c|}{\multirow{2}{*}{Datasets}}                              & \multicolumn{6}{c|}{Shiny   Blender}               & \multicolumn{6}{c}{Glossy Synthetic}            \\ \cline{3-14} 
\multicolumn{2}{c|}{}                                                       & ball  & car   & coffee & helmet & teapot & toaster & bell  & cat   & luyu  & potion & tbell & teapot \\ \midrule \midrule
\multicolumn{1}{c|}{\multirow{7}{*}{PSNR $\uparrow$}}      & Ref-NeRF~\cite{refnerf}       & 33.16 & \cellcolor{orange!30}30.44 & \cellcolor{yellow!30}33.99  & 29.94  & 45.12  & 26.12   & 30.02 & 29.76 & 25.42 & 30.11  & 26.91 & 22.77  \\
\multicolumn{1}{c|}{}                                      & NPC~\cite{npc}            & 23.76 & 24.19 & 30.39  & 25.59  & 41.22  & 19.76   & 22.41 & 25.35 & 23.68 & 23.09  & 19.03 & 18.21  \\
\multicolumn{1}{c|}{}                                      & 3DGS~\cite{kerbl2023gaussiansplatting}           & 27.65 & 27.26 & 32.3   & 28.22  & 45.71  & 20.99   & 25.11 & 31.36 & 26.97 & 30.16  & 23.88 & 21.51  \\
\multicolumn{1}{c|}{}                                      & GShader~\cite{jiang2024gaussianshader}        & 30.99 & 27.96 & 32.39  & 28.32  & \cellcolor{yellow!30}45.86  & \cellcolor{yellow!30}26.28   & 28.07 & \cellcolor{yellow!30}31.81 & 27.18 & 30.09  & 24.48 & 23.58  \\
\multicolumn{1}{c|}{}                                      & ENVIDR~\cite{liang2023envidr}         & \cellcolor{red!30}41.02 & 27.81 & 30.57  & \cellcolor{red!30}32.71  & 42.62  & 26.03   & \cellcolor{yellow!30}30.88 & 31.04 & \cellcolor{yellow!30}28.03 & \cellcolor{yellow!30}32.11  & \cellcolor{yellow!30}28.64 & \cellcolor{orange!30}26.77  \\

\multicolumn{1}{c|}{}                                      & 3DGS-DR~\cite{3dgsdr} & \cellcolor{yellow!30}33.66 & \cellcolor{yellow!30}30.39 & \cellcolor{orange!30}34.65  & \cellcolor{yellow!30}31.69  & \cellcolor{orange!30}47.12  & \cellcolor{orange!30}27.02   & \cellcolor{orange!30}31.65 & \cellcolor{orange!30}33.86 & \cellcolor{orange!30}28.71 & \cellcolor{orange!30}32.29  & \cellcolor{orange!30}28.94 & \cellcolor{yellow!30}25.36  \\

 \multicolumn{1}{c|}{}   &  3DGS-DR (\textbf{+Ours}) & \cellcolor{orange!30}34.51 & \cellcolor{red!30}30.83 &\cellcolor{red!30} 34.81  & \cellcolor{orange!30}32.24  & \cellcolor{red!30}47.93  &\cellcolor{red!30} 27.36   & \cellcolor{red!30}33.20 & \cellcolor{red!30}33.93 & \cellcolor{red!30}29.31 & \cellcolor{red!30}32.90  & \cellcolor{red!30}29.31 & \cellcolor{red!30}26.91 \\ \midrule

\multicolumn{1}{c|}{\multirow{7}{*}{SSIM $\uparrow$}}      & Ref-NeRF~\cite{refnerf}       & 0.971 & \cellcolor{yellow!30}0.950  & \cellcolor{orange!30}0.972  & 0.954  & 0.995  & 0.921   & 0.941 & 0.944 & 0.901 & 0.933  & \cellcolor{yellow!30}0.947 & 0.897  \\
\multicolumn{1}{c|}{}                                      & NPC~\cite{npc}            & 0.908 & 0.898 & 0.955  & 0.938  & 0.994  & 0.835   & 0.892 & 0.921 & 0.854 & 0.877  & 0.742 & 0.762  \\
\multicolumn{1}{c|}{}                                      & 3DGS~\cite{kerbl2023gaussiansplatting}           & 0.937 & 0.931 & \cellcolor{orange!30}0.972  & 0.951  & \cellcolor{yellow!30}0.996  & 0.894   & 0.908 & 0.959 & 0.916 & 0.938  & 0.900   & 0.881  \\
\multicolumn{1}{c|}{}                                      & GShader~\cite{jiang2024gaussianshader}        & 0.966 & 0.932 & \cellcolor{yellow!30}0.971  & 0.951  & \cellcolor{yellow!30}0.996  & \cellcolor{yellow!30}0.929   & 0.919 & 0.961 & 0.914 & 0.936  & 0.898 & 0.901  \\
\multicolumn{1}{c|}{}                                      & ENVIDR~\cite{liang2023envidr}         & \cellcolor{red!30}0.997 & 0.943 & 0.962  & \cellcolor{red!30}0.987  & 0.995  & 0.922   & \cellcolor{yellow!30}0.954 & \cellcolor{yellow!30}0.965 & \cellcolor{yellow!30}0.931 & \cellcolor{orange!30}0.960   & \cellcolor{yellow!30}0.947 & \cellcolor{red!30}0.957  \\
\multicolumn{1}{c|}{}                                      & 3DGS-DR~\cite{3dgsdr} & \cellcolor{yellow!30}0.979 & \cellcolor{orange!30}0.962 & \cellcolor{red!30}0.976  & \cellcolor{yellow!30}0.971  & \cellcolor{orange!30}0.997  & \cellcolor{orange!30}0.943   & \cellcolor{orange!30}0.962 & \cellcolor{orange!30}0.976 & \cellcolor{orange!30}0.936 & \cellcolor{yellow!30}0.957  & \cellcolor{orange!30}0.952 & \cellcolor{yellow!30}0.936  \\ 

\multicolumn{1}{c|}{}                                      & 3DGS-DR (\textbf{+Ours}) & \cellcolor{orange!30}0.983 & \cellcolor{red!30}0.965 & \cellcolor{red!30}0.976  & \cellcolor{orange!30}0.974  & \cellcolor{red!30}0.998  & \cellcolor{red!30}0.949   & \cellcolor{red!30}0.974 & \cellcolor{red!30}0.979 & \cellcolor{red!30}0.947 & \cellcolor{red!30}0.963  & \cellcolor{red!30}0.965 & \cellcolor{orange!30}0.942  \\ \midrule

\multicolumn{1}{c|}{\multirow{7}{*}{LPIPS   $\downarrow$}} & Ref-NeRF~\cite{refnerf}       & 0.166 & 0.050  & 0.082  & 0.086  & 0.012  & 0.083   & 0.102 & 0.104 & 0.098 & 0.084  & 0.114 & 0.098  \\
\multicolumn{1}{c|}{}                                      & NPC~\cite{npc}            & 0.237 & 0.120  & 0.119  & 0.156  & 0.013  & 0.226   & 0.203 & 0.121 & 0.101 & 0.174  & 0.243 & 0.246  \\
\multicolumn{1}{c|}{}                                      & 3DGS~\cite{kerbl2023gaussiansplatting}           & 0.162 & 0.047 & 0.079  & 0.081  & 0.008  & 0.125   & 0.104 & 0.062 & 0.064 & 0.093  & 0.125 & 0.102  \\
\multicolumn{1}{c|}{}                                      & GShader~\cite{jiang2024gaussianshader}        & 0.121 & \cellcolor{yellow!30}0.044 & \cellcolor{yellow!30}0.078  & 0.074  & \cellcolor{yellow!30}0.007  & \cellcolor{orange!30}0.079   & 0.098 & 0.056 & 0.064 & 0.088  & 0.122 & 0.091  \\
\multicolumn{1}{c|}{}                                      & ENVIDR~\cite{liang2023envidr}         & \cellcolor{red!30}0.020  & 0.046 & 0.083  & \cellcolor{red!30}0.036  & 0.009  & \cellcolor{yellow!30}0.081   & \cellcolor{yellow!30}0.054 & \cellcolor{yellow!30}0.049 & \cellcolor{yellow!30}0.059 & \cellcolor{orange!30}0.072  & \cellcolor{yellow!30}0.069 & \cellcolor{red!30}0.041  \\

\multicolumn{1}{c|}{}                                      &    3DGS-DR~\cite{3dgsdr}            & \cellcolor{orange!30}0.098 & \cellcolor{orange!30}0.033 & \cellcolor{orange!30}0.076  & \cellcolor{yellow!30}0.049  & \cellcolor{orange!30}0.005  & \cellcolor{yellow!30}0.081   & \cellcolor{orange!30}0.046 & \cellcolor{orange!30}0.040  & \cellcolor{orange!30}0.053 & \cellcolor{yellow!30}0.075  & \cellcolor{orange!30}0.067 & \cellcolor{yellow!30}0.067  \\

\multicolumn{1}{c|}{}                                      &   3DGS-DR (\textbf{+Ours})             & \cellcolor{yellow!30}0.089 & \cellcolor{red!30}0.030 & \cellcolor{red!30}0.074  & \cellcolor{orange!30}0.042  & \cellcolor{red!30}0.004  & \cellcolor{red!30}0.067   & \cellcolor{red!30}0.031 & \cellcolor{red!30}0.035  & \cellcolor{red!30}0.044 & \cellcolor{red!30}0.062  & \cellcolor{red!30}0.048 & \cellcolor{orange!30}0.060  \\ \bottomrule
\end{tabular}
\end{adjustbox}
\label{t_reflect}

\end{table*}

\begin{figure*}[t!]
    \centering
    \includegraphics[width=0.99 \linewidth]{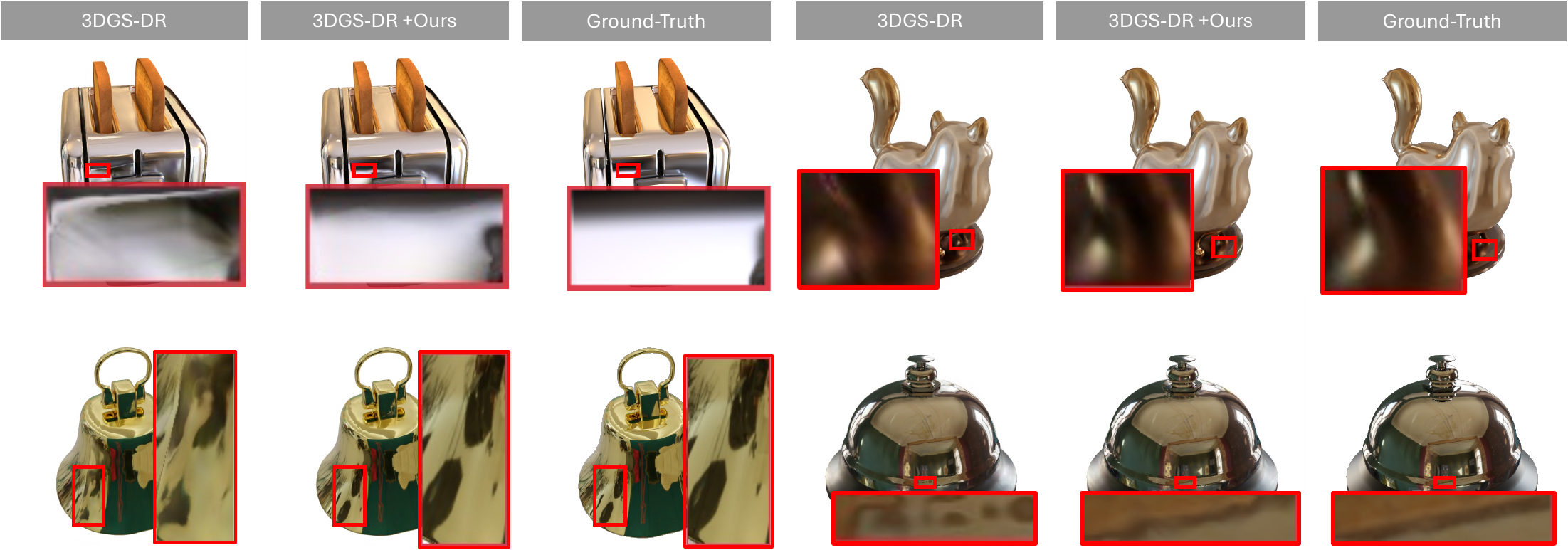} 
    \caption{ \textbf{Qualitative results of 3DGS-DR~\cite{3dgsdr} and its improved version by integrating our method across various challenging datasets.} It can be observed that 3DGS-DR integrated with our method can achieve better results for extremely challenging scenes with strong reflection.}
    \label{vis_reflective}
\end{figure*}

\subsection{Reflective Object Novel View Synthesis}
To demonstrate the generalization of our proposed method, we conduct experiments for the reflective object Novel View Synthesis task.
In particular, this task is more challenging than generic object NVS because it contains objects with strong reflections and drastic lighting effect variation. 
As depicted in Table~\ref{t_reflect}, we compare several state-of-the-art reflective object NVS methods,
including Ref-NeRF~\cite{refnerf}, NPC~\cite{npc}, 3DGS~\cite{kerbl2023gaussiansplatting}, GaussianShader~\cite{jiang2024gaussianshader}, ENVIDR~\cite{liang2023envidr}, 3DGS-DR~\cite{3dgsdr}, and our improved version on 3DGS-DR. 
Specifically, we conduct experiments on two commonly used public datasets, like Shiny Blender~\cite{refnerf} and Glossy Synthetic dataset~\cite{liu2023nero}. 
In Table~\ref{t_reflect}, it can be observed that our method integrated into 3DGS-DR achieves superior quantitative results compared with existing methods. In addition, we also present visual comparisons in Fig.~\ref{vis_reflective} to assess our method qualitatively. We found that 3DGS-DR cannot accurately recover lighting effects on glossy surfaces and fine details reflecting the surrounding environments. 
In particular, the reconstruction of specular lighting effects is challenging since the specular effects are often view-dependent. In other words, specular occurs when light reflects off a surface in a specific direction, and the appearance depends on the position of the light source and the observer.
We attribute the improvements to the proposed multi-view regulated learning strategy. Rather than relying on single-view constraints alone, our method incorporates multi-view constraints that guide the optimization of Gaussian attributes, promoting a more coherent representation of specular across viewpoints. Additionally, this strategy encourages more effective densification for 3D Gaussians, enabling finer modeling of specular and surrounding reflections.


\begin{table*}[t]
\caption{\textbf{Quantitative comparisons of state-of-the-art multi-scale scene reconstruction methods.} We demonstrate our method can also improve novel view synthesis performance for challenging multi-scale scenes. We report results on BungeeNeRF datasets~\cite{xiangli2022bungeenerf}.}
\centering
\def\arraystretch{1.2}

\resizebox{0.95\linewidth}{!}
{
    \begin{tabular}{l | ccc | ccc | ccc}
    \toprule
    Scene & \multicolumn{3}{c|}{Chicago} & \multicolumn{3}{c|}{Rome} & \multicolumn{3}{c}{Hollywood} \\
    \midrule \midrule 
    Method \& Metrics & PSNR$\uparrow$ & SSIM$\uparrow$ & LPIPS$\downarrow$ & PSNR$\uparrow$ & SSIM$\uparrow$ & LPIPS$\downarrow$ & PSNR$\uparrow$ & SSIM$\uparrow$ & LPIPS$\downarrow$ \\
      \midrule 
    3DGS~\cite{kerbl2023gaussiansplatting} & 28.17 & \cellcolor{yellow!30}0.930 & 0.084 & 27.54 & 0.916 & 0.100 & 26.24 & 0.869 & 0.133 \\
    Mip-Splatting~\cite{mipsplatting} & 28.28 & \cellcolor{yellow!30}0.930 & 0.081 & \cellcolor{yellow!30}28.33 & 0.922 & 0.093 & \cellcolor{yellow!30}26.59 & \cellcolor{yellow!30}0.876 &  \cellcolor{yellow!30}0.130 \\ 
    Scaffold-GS~\cite{lu2024scaffold}& \cellcolor{yellow!30}28.55 & 0.929 & \cellcolor{yellow!30}0.080 & 28.24 & \cellcolor{yellow!30}0.924 & \cellcolor{yellow!30}0.087 & 26.36 & 0.866 & 0.157 \\
    
    Octree-GS~\cite{ren2024octree} & \cellcolor{orange!30}28.62 & \cellcolor{orange!30}0.934 & \cellcolor{orange!30}0.075 & \cellcolor{orange!30}28.50 & \cellcolor{orange!30}0.932 & \cellcolor{orange!30}0.077 & \cellcolor{orange!30}26.70 & \cellcolor{orange!30}0.885 & \cellcolor{orange!30}0.126 \\
    \midrule
    Octree-GS (+Ours) & \cellcolor{red!30}28.82 &  \cellcolor{red!30}0.936 & \cellcolor{red!30}0.069 &  \cellcolor{red!30}28.79 & \cellcolor{red!30}0.933 & \cellcolor{red!30}0.073 & \cellcolor{red!30}26.73 & \cellcolor{red!30}0.887 & \cellcolor{red!30}0.122\\
    
    \bottomrule
    \end{tabular}
}
\label{t_bungeenerf}
\end{table*}

\begin{figure*}[t!]
    \centering
    \includegraphics[width=1 \linewidth]{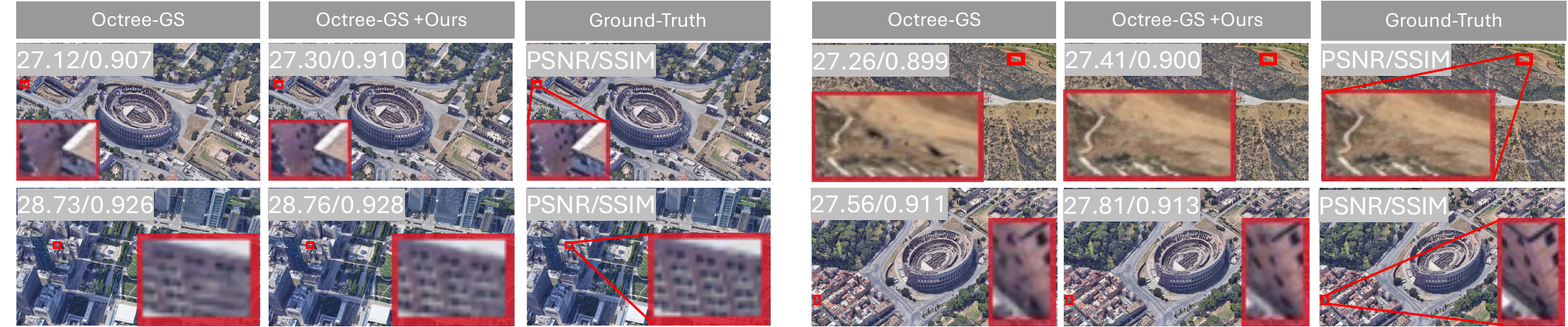} 
    \caption{\textbf{Qualitative comparisons of multi-scale scene Novel View Synthesis on BungeeNeRF dataset~\cite{xiangli2022bungeenerf}. We compare Octree-GS~\cite{ren2024octree} and its improved version by integrating our method.} We utilize \textcolor{red}{red} close-up patches to differentiate the visual differences for clear comparisons. It is observed that our proposed method can improve the original Octree-GS for challenging multi-scale scenes. }
    \label{vis_bungee}
\end{figure*}

\subsection{Large-scale Scene Novel View Synthesis} We additionally conduct experiments on a large-scale scene dataset, BungeeNeRF~\cite{xiangli2022bungeenerf}, to further prove the effectiveness of our method.  As depicted in Table~\ref{t_bungeenerf}, we report results on three representative scenes with existing state-of-the-art methods, including 3DGS~\cite{kerbl2023gaussiansplatting}, Mip-Splatting~\cite{mipsplatting}, Scaffold-GS~\cite{lu2024scaffold},  Octree-GS~\cite{ren2024octree}, and our improved version of Octree-GS.
Table~\ref{t_bungeenerf} demonstrates that our proposed method improves the recent state-of-the-art Octree-GS for better novel view synthesis results. 
This improvement is due to the proposed multi-view training and densification strategies, constraining with multi-view supervision and producing more 3D Gaussians for faster convergence and finer detailed novel views.
These results also imply that our method is able to generalize to diverse scenes, although they are not object-centered. 

We also provide the visualization results of multi-scale scene Novel View Synthesis on the BungeeNeRF dataset~\cite{xiangli2022bungeenerf} in Fig.~\ref{vis_bungee}. We observe that the original Octree-GS~\cite{ren2024octree} struggles to reconstruct intricate details and tends to produce noticeable artifacts, which decrease its overall Novel View Synthesis quality. In contrast, the enhanced version of Octree-GS, empowered by our proposed method, successfully captures and renders finer texture details, closely resembling the ground truth. This improvement underscores the robustness and precision of our approach, as it significantly reduces artifacts while enhancing visual quality across complex scenes.

These experiments not only highlight the effectiveness of our method but also demonstrate its generalization to a wide variety of scenes, even for complex environments. These results indicate the applicability of our approach across different Novel View Synthesis tasks. Moreover, the results also suggest that our method consistently improves rendering quality, particularly for novel view synthesis. Its versatility extends across multiple applications such as general object NVS, reflective object NVS, 4D dynamic scene NVS, and multi-scale scene NVS. These findings emphasize the broad potential of our approach in advancing the state-of-the-art Gaussian-based methods for novel view synthesis.

\begin{table*}[t!]
\centering  
\scriptsize
\def\arraystretch{1.2}
\resizebox{\linewidth}{!}{
\begin{threeparttable}
\caption{\textbf{Per-scene quantitative results for 4D reconstruction on the D-NeRF~\cite{dnerf} dataset. } 
We integrate our method into 4DGS and improve its 4D reconstruction performance.  }
\begin{tabular}{l|ccc|ccc|ccc|ccc}  
    \toprule  
    \multirow{2}{*}{Method}&  
    \multicolumn{3}{c}{Bouncing Balls}&\multicolumn{3}{c}{Hellwarrior}&\multicolumn{3}{c}{Hook}&\multicolumn{3}{c}{Jumpingjacks}\cr  
    \cmidrule(lr){2-4}\cmidrule(lr){5-7}\cmidrule(lr){8-10}\cmidrule(lr){11-13}
    &PSNR&SSIM&LPIPS&PSNR&SSIM&LPIPS&PSNR&SSIM&LPIPS&PSNR&SSIM&LPIPS\cr  
    \midrule  
    3DGS~\cite{kerbl2023gaussiansplatting}&23.20&0.959&0.060&24.53&0.933&0.058\cellcolor{yellow!30}&21.71 & 0.887& 0.103& 23.20&0.959 &0.060\cr  
    K-Planes~\cite{kplane}&40.05&\cellcolor{yellow!30}0.993&0.032\cellcolor{yellow!30}&24.58&0.952&0.082&28.12&0.948&0.066&31.11&0.970&0.046\cr  
    HexPlane~\cite{cao2023hexplane}&39.86&0.991&0.032&24.55&0.944&0.073&28.63&0.957\cellcolor{yellow!30}&0.050\cellcolor{yellow!30}&31.31&0.972&0.039\cr  
    TiNeuVox~\cite{tineuvox}&\cellcolor{yellow!30}40.23&0.992&0.041&\cellcolor{yellow!30}27.10&\cellcolor{yellow!30}0.963&0.076&28.63\cellcolor{yellow!30}&0.943&0.063&33.49\cellcolor{yellow!30}&0.977\cellcolor{yellow!30}&0.040\cellcolor{yellow!30}\cr  	
    4DGS~\cite{4dgs}&\cellcolor{orange!30}40.62&\cellcolor{orange!30}0.994&\cellcolor{orange!30}0.015&\cellcolor{orange!30}28.71&\cellcolor{orange!30}0.973&\cellcolor{orange!30}0.036&\cellcolor{orange!30}32.73&\cellcolor{orange!30}0.976&\cellcolor{orange!30}0.027&\cellcolor{orange!30}35.42&\cellcolor{orange!30}0.985&\cellcolor{orange!30}0.012\cr

    4DGS + (\textbf{Ours})&\cellcolor{red!30}41.60&\cellcolor{red!30}0.995&\cellcolor{red!30}0.011&\cellcolor{red!30}29.29&\cellcolor{red!30}0.976&\cellcolor{red!30}0.029&\cellcolor{red!30}33.67&\cellcolor{red!30}0.979&\cellcolor{red!30}0.021&\cellcolor{red!30}37.69&\cellcolor{red!30}0.990&\cellcolor{red!30}0.011\cr  
\midrule \midrule
\multirow{2}{*}{Method}&  
    \multicolumn{3}{c}{Lego}&\multicolumn{3}{c}{Mutant}&\multicolumn{3}{c}{ Standup}&\multicolumn{3}{c}{Trex}\cr  
    \cmidrule(lr){2-4}\cmidrule(lr){5-7}\cmidrule(lr){8-10}\cmidrule(lr){11-13}
    &PSNR&SSIM&LPIPS&PSNR&SSIM&LPIPS&PSNR&SSIM&LPIPS&PSNR&SSIM&LPIPS\cr  
    \midrule  3DGS~\cite{kerbl2023gaussiansplatting}&23.06&0.929&0.064&20.64&0.929&0.082&21.91&0.930&0.078&21.93&0.953&0.048\cr  
    K-Planes~\cite{kplane}&\cellcolor{red!30}25.49&\cellcolor{red!30}0.948&\cellcolor{red!30}0.033&32.50&0.971&0.036&33.10&0.979&0.031&30.43&\cellcolor{yellow!30}0.973&0.034\cr  
    HexPlane~\cite{cao2023hexplane}&\cellcolor{orange!30}25.10&\cellcolor{orange!30}0.938&\cellcolor{yellow!30}0.043&33.67\cellcolor{yellow!30}&0.980\cellcolor{yellow!30}2&0.026\cellcolor{yellow!30}&34.40&\cellcolor{yellow!30}0.983&\cellcolor{yellow!30}0.020&30.67&\cellcolor{orange!30}0.974&\cellcolor{yellow!30}0.027\cr  
    TiNeuVox~\cite{tineuvox}&24.65&0.906&0.064&30.87&0.960&0.047&\cellcolor{yellow!30}34.61&0.979&0.032&\cellcolor{yellow!30}31.25&0.966&0.047\cr  
    4DGS~\cite{4dgs}&\cellcolor{yellow!30}25.03&\cellcolor{yellow!30}0.937&0.038\cellcolor{orange!30}&\cellcolor{orange!30}37.59&\cellcolor{orange!30}0.988&\cellcolor{orange!30}0.016&\cellcolor{orange!30}38.11&\cellcolor{orange!30}0.989&\cellcolor{red!30}0.007&\cellcolor{orange!30}34.23&\cellcolor{red!30}0.985&\cellcolor{red!30}0.013\cr  

4DGS (\textbf{+Ours})&24.70&0.932&0.057&\cellcolor{red!30}38.82&\cellcolor{red!30}0.991&\cellcolor{red!30}0.012&\cellcolor{red!30}40.81&\cellcolor{red!30}0.993&\cellcolor{orange!30}0.008&\cellcolor{red!30}34.26&\cellcolor{red!30}0.985&\cellcolor{orange!30}0.019\cr  
    \bottomrule  
\end{tabular}  
\label{t_4d}
\end{threeparttable}
}
\end{table*}

\begin{figure}[t!]
    \centering
    \includegraphics[width=0.95 \linewidth]{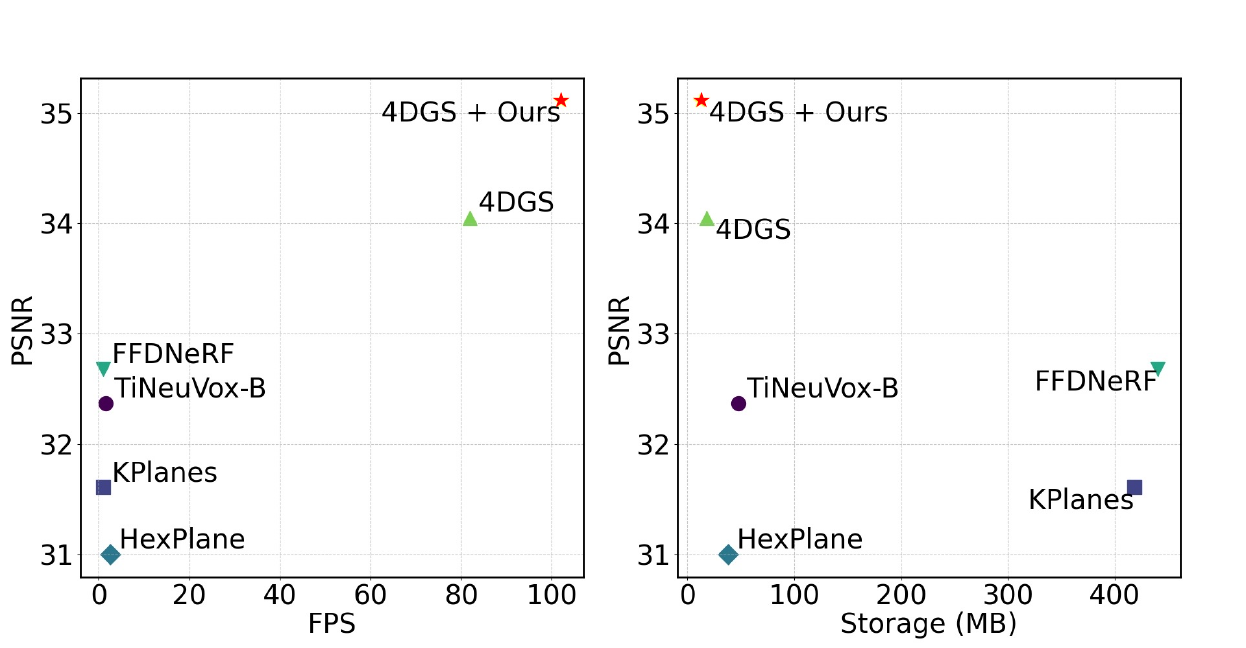} 
    \caption{ \textbf{FPS and storage memory comparisons between existing state-of-the-art 4D Novel View Synthesis methods and our proposed method.} We demonstrate that our proposed method integrated into 4DGS~\cite{4dgs} achieves the best PSNR results with the fastest FPS rendering speed and the smallest storage memory. }
    \label{vis_4d_fps}
\end{figure}

\begin{figure*}[t!]
    \centering
    \includegraphics[width=1 \linewidth]{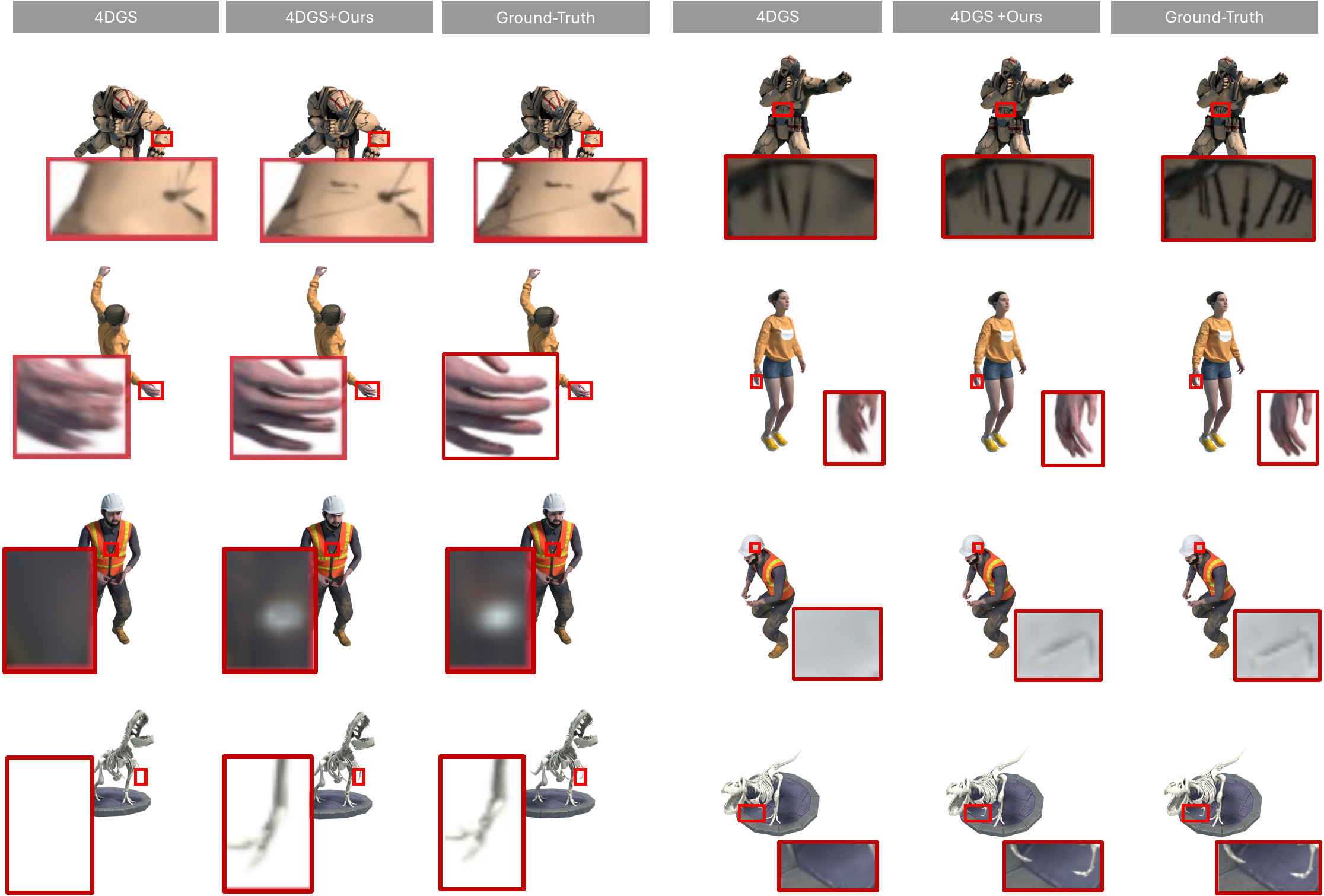} 
   \caption{\textbf{Visualization comparisons of 4DGS~\cite{4dgs} and its improved version integrating with our method across various dynamic scenes.}
    Our proposed method can impose the original 4DGS with multi-view constraints for better dynamic scene Novel View Synthesis performance with finer details.
        We scale up red patches for clearer visibility.
    }
    \label{vis_4d}
\end{figure*}

\subsection{4D Novel View Synthesis}
To further demonstrate the effectiveness of our proposed method, we conduct experiments for the task of 4D Novel View Synthesis. 
4D Novel View Synthesis is more challenging than 3D Novel View Synthesis since it contains the dimension of time, and the scenes change over time. 
In Table \ref{t_4d}, we present detailed per-scene quantitative results on the D-NeRF~\cite{dnerf} dataset for the evaluation of 4D Novel View Synthesis performance across state-of-the-art methods, including 3DGS~\cite{kerbl2023gaussiansplatting}, K-Planes~\cite{kplane}, HexPlane~\cite{cao2023hexplane}, TiNeuVox~\cite{tineuvox}, 4DGS~\cite{4dgs}, and our improved version of 4DGS by integrating with our method.
It can be observed that our method integrated into 4DGS~\cite{4dgs} achieves state-of-the-art results compared with existing 4D Novel View Synthesis methods. 
In addition, we also report the rendering speed (FPS) and storage size (MB) metrics as shown in Fig.~\ref{vis_4d_fps}. 
We observed that 4DGS integrated with our method achieves faster rendering speed with fewer 4D Gaussian parameters.  
This is because our proposed multi-view training provides the holistic constraints toward 4D Gaussian structures, enabling the structure to be more compact in temporal and spatial dimensions.
Therefore, it leads to fewer 4D Gaussian parameters and faster rendering speed.

In addition to the quantitative analysis, we also present visual comparisons in Fig. \ref{vis_4d} to further evaluate the qualitative performance of our method. As depicted in Fig. \ref{vis_4d}, 4DGS struggles to reconstruct fine details, often failing to capture subtle textures and intricate structural elements. In contrast, our approach, when integrated into 4DGS, yields a substantial improvement, enabling the Novel View Synthesis of much finer details with clearer and more accurate texture representation. 
The key lies in our proposed MVGS, which leverages multi-view supervision across both spatial and temporal domains. By jointly optimizing the 4DGS over multiple views and time steps, our method enforces consistency and coherence, encouraging the reconstruction of better structures. This multi-view constraint acts as a powerful regularizer, enabling the system to recover high-frequency details and maintain structural accuracy even in challenging dynamic scenarios.
These visual results, as well as the quantitative improvement, demonstrate that our method not only enhances the richness and sharpness of the reconstructed scenes but also significantly reduces artifacts, leading to a more realistic and lifelike representation of 4D dynamic scenes.  It is attributed to our proposed multi-view training method that constrains the optimization of 3D Gaussians with multi-view information, especially in dynamic scenes with temporally varying views. 

The qualitative and quantitative experimental results strongly demonstrate the superiority of our method over existing approaches. Our method consistently leads to performance improvement across a wide range of tasks, proving its versatility and robustness in enhancing 4D Novel View Synthesis. Furthermore, we demonstrate that our approach can be adapted to different scenes and capture intricate details in dynamic scenes, showcasing its potential for a broad spectrum of applications such as motion capture, virtual reality, and high-fidelity simulations. The results further consolidate the contribution of our method in improving the state-of-the-art 4D Novel View Synthesis.


\bibliographystyle{splncs04}
\bibliography{main}

\end{document}